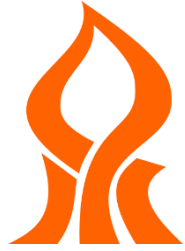

**Ben-Gurion University of the Negev**
**The Faculty of Engineering Sciences**
Department of Industrial Engineering and Management

# Mobile robot's sampling algorithms for monitoring of insects' populations in agricultural fields

By: Adi Yehoshua

Supervised by: Prof. Yael Edan

March 2023

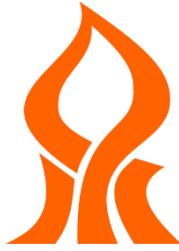

Ben-Gurion University of the Negev
The Faculty of Engineering Sciences
Department of Industrial Engineering and Management

# Mobile robot's sampling algorithms for monitoring of insects' populations in agricultural fields

Thesis submitted in partial fulfillment of the requirements for the M.Sc. degree

By: Adi Yehoshua

Supervised by: Prof. Yael Edan

Author:…………………….………………  Date: 03.05.2023

Supervisor:…………….…………………..  Date: 03.05.2023

Chairman of Graduate Studies Committee:…………... …    Date:

March 2023

# Acknowledgments

To begin with, I am deeply appreciative of Prof. Yael Edan for being an outstanding supervisor who showed unwavering dedication, care, and professionalism. Through your guidance, I was exposed to new realms, and you served as a valuable mentor throughout my journey. You provided invaluable advice on research topics, connected me with necessary resources, and were always available and prompt in answering my every query. I am grateful for the open dialogue we shared, and I admire you as a role model. Your encouragement and support propelled me to succeed, and I am indebted to you for everything.

Additionally, I would like to express my gratitude to my colleagues, Avital Bechar, Yafit Cohen, Tohar Brener, and Liran Shmuel, for their support during our joint work on these two research projects. I cannot thank you enough for the assistance you provided me over the past few years.

Lastly, I would like to extend my heartfelt appreciation to my supportive family and friends who accompanied me, listened to me, and motivated me throughout this research endeavor.

# Abstract

Pests and plant diseases are major causes of production losses and may have a significant impact on the agricultural sector. Detecting pests as early as possible can help increase crop yields and production efficiency. Several robotic monitoring systems have been developed allowing to collect data and provide a greater understanding of environmental processes. An agricultural robot can enable accurate timely detection of pests, by traversing the field autonomously and monitoring the entire cropped area within a field. However, in many cases it is impossible to sample all plants due to resource limitations, such as financial, temporal, or energy constraints.

In this thesis, the development and evaluation of several sampling algorithms is presented to address the challenge of an agriculture-monitoring ground robot designed to locate insects in an agricultural field, where complete sampling of all the plants is infeasible due to resource constraints.

Two situations were investigated in simulation models that were specially developed as part of this thesis: a situation where no a-priori information on the insects is available and a situation where prior information of the insects distributions within the field are known.

For the situation of no a-priori information on the field and/or the insects in it, seven algorithms were devised and tested, each utilizing a distinct approach to sample the field without prior knowledge on it. The results of the research yielded valuable insights. For small-sized fields and a single working robot, thorough sampling is crucial and skipping plants should be kept to a minimum. The best detection rates were observed with algorithms that visit every plant, and skipping plants led to decreased success rates. For larger fields, the algorithm that implements the neighbor strategy, which provides thorough coverage without skipping any plants and allows for control of a larger proportion of the field each day, is preferred. This algorithm has the highest potential to reach 100% detection, regardless of field size.

For the situation of having prior information on the insects locations in the field, we present the development and evaluation of a dynamic sampling algorithm The algorithm utilizes real-time information to prioritize sampling at suspected points, locate hot spots and adapt sampling plans accordingly. The algorithm's performance was compared to two existing algorithms using *Tetranychidae* insect data from previous research. Analyses revealed that the dynamic algorithm outperformed the others in all tested scenarios, reaching 100% detection approximately 3–5 days sooner when applied to small fields, and identifying 30%–50% more insects for larger fields. Its high detection percentages in small fields—100 for a 1 ha field—decreased moderately with increasing field size to 80% for a 10 ha field, seemingly irrespective of insect spread rate, which also barely affected insect detection. Doubling the time spent on each sample improved the results by 30–50% on average in the early stages of the simulation.

# Publications

Adi Yehoshua, Avital Bechar, Yafit Cohen, Liran Shmuel, Yael Edan.

Dynamic sampling algorithm for agriculture-monitoring ground robot.

Submitted February 2023. IJSIMM 646-2023.

# Table of Contents

# List of Figures

# List of Tables

# 1. Introduction

## 1.1. Overview and problem description

The impending global demand for food is projected to rise by 60%, with an increase of 100% in developing nations (Maxted et al., 2016). This is an outcome of the combined pressures of population growth and a shift towards more resource-intensive dietary habits (Maxted et al., 2016). To meet this growing demand, it is imperative to augment crop yields and production efficiency (Muangprathub et al., 2019). Pests and plant diseases are a major cause of production losses and economic impacts in the agricultural sector (Savary et al., 2012). Direct yield losses caused by pathogens are estimated to range between 20-40% of global agricultural productivity (Savary et al., 2012). Thus, early detection and management of pests is crucial in order to mitigate negative effects on the quality and quantity of agricultural produce (Savary et al., 2012). In order to improve agriculture production and efficiency, the field of agricultural robotics has seen extensive research and development for a variety of field operations (Roldan et al., 2018). Robotic systems can perform tasks such as planting, harvesting, and monitoring crop health, freeing up human workers for more complex tasks (Fountas et al., 2020), while improving crop yields and reducing waste (Marinoudi et al., 2019). Robotic technology can assist in early detection and management of pests and plant diseases by applying sensors (Li et al., 2010), machine learning algorithms (Liakos et al., 2018), and precision agriculture techniques (Marinoudi et al., 2019). An agricultural robot can efficiently cover the entire cropped area within a field by autonomously traversing the field (Hameed et al., 2019). However, due to resource limitations, such as financial, temporal, or energy constraints, in many cases the robotic system cannot traverse the whole agricultural field. In these cases the implementation of an appropriate sampling algorithm becomes essential. (Hameed et al., 2013). This algorithm can allow for effective and efficient sampling of the plants in the field overcoming the limitations.

However, only a small amount of research has focused on sampling algorithms (Bergman et al., 2016). Furthermore, there is a lack of research on improving cycle time in pest inspection tasks (Schor et al., 2017), which encompass path planning, navigation, sensing, mapping, and action operations (Botta et al., 2022). A primary bottleneck in agricultural monitoring robots is the time required for travel between plants (Santos et al., 2020). The most common practice for carrying out tasks such as spraying, detecting insects, and monitoring is to traverse the entire field and sample all plants, without distinguishing between them.

A wealth of research studies have aimed at finding efficient routes for field coverage planning, with the goal of minimizing travel time (Hameed et al., 2013). A variety of algorithms have been applied, including greedy algorithms (Oksanen & Visala, 2009), the Dragonfly algorithm (Muthukumaran & Sivaramakrishnan, 2019), genetic algorithms (Burchardt & salomon, 2006; Manikas et al., 2007), ant colony optimization (Akka & khaber, 2018), A* and dynamic A* (D*) (Guo et al., 2009). All these algorithms aim to determine the optimal route for navigation in the field- by using prior knowledge if it is available, or by preliminary mapping the area if there is no prior information on it. Yet none suggest which plants should be sampled and in what order. Due to limitations of resources such as time and energy, it is often not feasible for the robot to sample all the plants field. In such scenarios, a sampling algorithm is necessary to prioritize and determine which plants should be sampled. This algorithm can help improve resource usage and ensure that the most important plants are sampled.

Previous studies have revealed that the dissemination of pests exhibits a fixed pattern (Shmuel, 2018), wherein the mode of entry into the field and the manner in which they propagate within it is determined. Pests tend to primarily spread from the boundaries of the field and the plants that are situated adjacent to paths within the field (Shmuel, 2018). Conversely, plants located in the interior of the field are less likely to initiate the spread. Another insight gleaned from prior research is the trajectory of the pest's spread within the field over time (Cieniewicz et al., 2017). The denser the field (i.e., the closer proximity of plants to one another), the greater the likelihood for pests to spread to nearby plants and create hot spots (areas with high concentrations of pests) within the field. Conversely, in more spacious fields, the spread of pests is slower (Cieniewicz et al., 2017). A dynamic algorithm for determining which plants will be sampled and in what order can assist in situations where it is not feasible to sample the entire field due to resource constraints and given the understanding of pest dissemination patterns. This can ensure that the limited resources are directed towards the areas of highest risk, thus maximizing the effectiveness of insect management efforts.

## 1.2. Research Objectives

The primary objective of this thesis was to develop sampling algorithms for a mobile robot monitoring insects populations in agricultural fields.

The specific objectives were to:

1. Develop sampling algorithms to efficiently locate insects in an agricultural field.
2. Evaluate the performance of the proposed algorithms and compare them to the current state-of-the-art techniques.
3. Provide a comprehensive evaluation of the proposed algorithms in terms of its effectiveness, efficiency, and robustness in various scenarios.

# 2. Literature review

## 2.1. Robotics in agriculture

Robotics R&D have intensively been developed and include many applications such as harvesting of field crops (Hoffman et al., 1996), fruit picking in an orchard, a semi-autonomous robot for picking oranges (Hannan et al., 2004, Edan et al., 2000) cabbage (Reina et al., 2006), cherries (Imagawa et al., 2008), greenhouse robots for harvesting tomatoes (Yoshhiko et al., 2001), cucumbers (Van Henten et al., 2002), sweet and hot peppers (Kurtser et al. 2017, Kitamura, 2005). Despite significant advance in robotics, agriculture is still limited in applications due to the high variability both in the crop and environment (Kurster; Edan, 2018) along with the need for relatively low systems costs and the requirement for systems that can perform robustly and be operated by non-professional users (Edan, 2019). Commercial applications are penetrating into the market but still lack wide adoption and are limited in the scope of application. An agricultural robot consists of the following basic modules: a sensing system to measure physical and biological properties of the agricultural system (crop, soil, environment); decision- making capabilities to process the sensory system information; and actuators to manipulate the agricultural system.

## 2.2. Robots for ecological monitoring

Robotic systems are increasingly being utilized as data-gathering tools, allowing new perspectives and a greater understanding of environmental processes (Dunbabin & Marques, 2012). Opposed to traditional sensors that provide fixed monitoring points, robots can adapt to changes in the surrounding environment and moreover can manipulate the sensors to the optimal pose or to interact with objects in the environment in order to collect quality data. Extensive research has been conducted on the application of marine, terrestrial, and airborne robotic systems to a variety of ecological monitoring tasks which include oceanographic measurements (Wadhams et al., 2006), geological and volcanos (Astuti et al., 2009), atmospheric conditions, dust and gasses monitoring (Reggente et al., 2010; Dunbabin & Grinham, 2010), mapping (Bryson et al., 2010), monitoring of physical properties of the environment and animal tracking (Former et al., 2010). Despite these tremendous efforts, to the best of our knowledge, the literature on development of robotic systems for insect detection and classification is sparse and focuses only on bees (Hart & Huang, 2011) or locust swarm tracking (Tahir & Brooker, 2009).

## 2.3. Ecological importance of insect monitoring

Ecological monitoring is designed to expand the knowledge of the main causes that are responsible for changes or threats in the ecosystem, and to identify and manage the various possible actions that are needed (Schaeffer et al., 1998; Gitzen et al, 2012). Monitoring is usually performed in natural or in agricultural environments by measuring ecosystem state variables in a definite space and time domain, through repeated observations. The collected data must be analyzed and understood in relation to the environment to derive the ecosystem health status, to identify possible threats and to make decisions for management strategies. Ecological monitoring of insects (and arthropods in general) involves inspection of habitats for collection of representative specimens, their rough separation for selecting the target groups, and counting and classification of relevant samples (Basset et al., 1998).

Insects monitoring is typically focused on specific functional groups which are indicators of the status of the ecosystem, or which are known to impact on it: key or invasive pests, generalist or key natural enemies (predator/parasitoid), parasites or key parasites, hyperparasites, spiders, decomposers etc. The oscillations in population of these functional groups depends on many different variables such as food availability, climate, environment, and ecological interactions (Herricks, & Cairns, 1982). Beside biotic data on abundance and diversity of the taxa detected (Supriatna, 2018), a more comprehensive vision of the ecosystem can be gained by recording the associated abiotic data, as e.g. micro-climate or other physical-chemical parameters of the habitat (Lupi et al. 2013). The experimental design is highly influential on the accuracy of the assessed composition and structure of insect communities (Ermakov 2013, Farinha et al. 2014). Very commonly, scientists must compromise between the amplitude of dataset to acquire and the resources and time needed to obtain it (spatial pattern of monitoring and time frequency, number of units monitored, plants monitored). The availability of automated solutions could greatly enlarge the acquired dataset, enabling a more detailed and comprehensive vision of the ecosystem monitored.

## 2.4. Insect monitoring techniques

Current techniques for insect populations monitoring are either static or dynamic (Smith & McKenzie, 2004). Static techniques are based on a variety of trapping devices (e.g., adhesive, sucking, colored, light-based, pitfall) which are placed in fixed sampling points and collect specimens relying on the movement of insects towards the trap. To amplify the trapping capacity, often an attractant (e.g., light, color, food, pheromone, odor) is added, providing an action that may interfere on the accuracy of population density data recorded. Each trap can sample a single point and a limited portion of the surrounding environment but, on the other hand, it operates in the field without involving humans, except for periodical collections of captured specimens for their subsequent classification. An advantage of static traps is the possibility of incorporating low cost imaging and transmitting technologies for automated acquisition and transfer of static images to a remote server, to be processed or visually inspected (Cho, 2007; Liu et al., 2009; Guarnieri et al., 2011; Tirelli et al., 2011, Schor et al., 2017), or including analysis capability to process high-speed images for extraction and transmission of more advanced information about attracted insects (Oberti et al., 2008).

State of the art techniques for dynamic (moving) monitoring relies on human experts who move around in the environment along specific patterns to collect specimens. Collecting is generally actuated by means of simple tools as nets, sticks and trays, pooter-suckers.

A common technique used to collect insects within vegetation and leaves is the frappage or beating tray method (Christieet al., 2010), which consists in gently knocking on a plant branch over a tray or an umbrella-net for collecting the insects dislodged from the canopy. The specimens in the net are then immediately visually inspected or they can be singularly collected and examined by a pooter, or similar suction apparatus, which convey the insects though a vacuum pipe into a vial (Dietrick et al., 1960; Buffington & Redak, 1998). There are some examples of devices introduced to assist the manual collection and reducing the needed time, while increasing the repeatability of the sampling procedure. These devices are based on brushing tools (Upton & Mantle, 2010; Macmillan, 2015) to remove material from leaves surface, or on vacuum tools (Gary & Marston, 1976; Motamedinia & Rakhshani, 2017) able to suck out insects from vegetation and convey them in a collecting jar. Even if assisted with these devices, dynamic monitoring techniques imply great efforts in terms of human resources, are time consuming, require to visit different sites in narrow time windows, need repetitions in different period of the year.

## 2.5. Path planning methods

A path planning problem is the derivation of a pre-determined continuous curve, or path, in a configuration space from the starting configuration of a mobile unit to the end configuration, without having the mobile unit colliding with any obstacles (Choset, 2005). The last decades, because of the introduction of new autonomous and robotic systems there is an increasing demand for path planning approaches and implementations. Implementations of automated path planning include indoor implementations, e.g. for indoor transportation robots in hospitals (Takahashi, Suzuki, Shitamoto, Moriguchi, & Yoshida, 2010), industrial robot manipulation (Ting, Lei, & Jar, 2002), robots moving goods in warehouses (Wurman & Andrea, 2008) and optimal trajectory planning for surface spray painting (Diao, Zeng, & Tam, 2009). Outdoor implementations, include the task of following roads and parking among other moving cars (Likhachev & Ferguson, 2009), loading/ unloading of ship container crane (Huang, Liang, & Yang, 2009), shortest-path databases for parcel delivery systems (Jung, Lee, & Chun, 2006), routing Personal Rapid Transit vehicles in a rail network (Berger et al., 2011) There are four basic components of this process as shown in Figure 2.

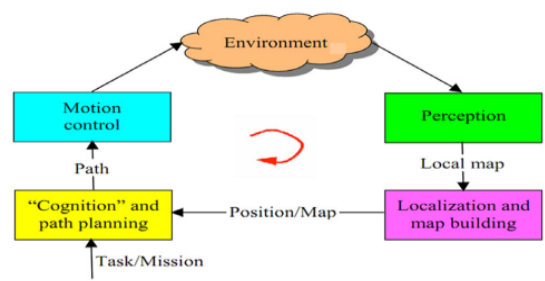

*Figure 1: Robot path planning structure (Mac et al., 2016)*

(i) perception, the robot uses its sensors to extract meaningful information.

(ii) localization, the robot determines its location in the working space.

(iii) cognition and path planning, the robot decides how to steer to achieve its goal.

(iv) motion control, the robot regulates its motion to accomplish the desired trajectory.

Path planning of a robot can be considered as a sequence of translation and rotation movements from a starting position to a destination while avoiding obstacles in its working environment. There are two suggested techniques in robot path planning:

(i) global path planning or off-line path planning (Qin, Sun, Li, Cen, 2017).

(ii) local path planning or on-line path planning (Raja, Pugazhenthi, 2012).

A global path planner usually generates a low-resolution high-level path based on a known environmental map or its current and past perceptive information of the environment. The method is valuable of producing an optimized path; however, it is inadequate reacting to unknown or dynamic obstacles. Local path planning algorithms do not need a priori information of the environment. It usually gives a high-resolution low-level path only over a fragment of global path based on information from on-board sensors. It works effectively in dynamic environments. The method is inefficient when the target is long distance away or the environment is cluttered. Normally, the combination of both methods is advised to enhance their advantages and eliminates some of their weaknesses (Zhang, Butzke, Likhachev, 2012). The robot path planning problem can be divided into classical methods and heuristic methods (Masehian, Sedighizadeh & Mohanty, Parhi, 2013), as shown in Figure 3.

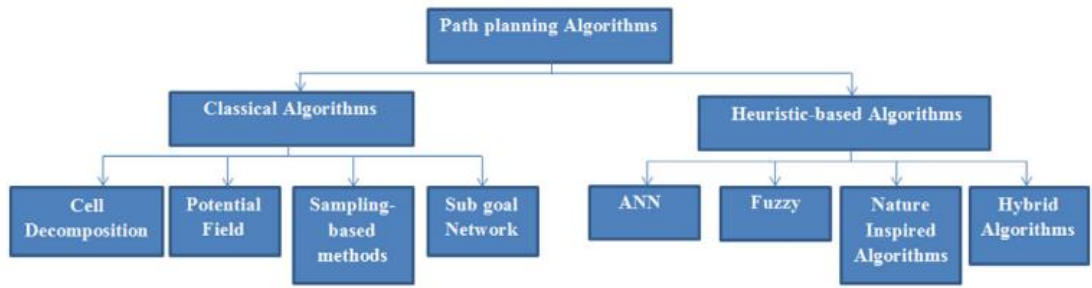
*Figure 2: Path planning algorithms (Mac et al., 2016)*

## 2.6. Field coverage planning algorithms

A wealth of research studies have aimed at finding efficient routes for field coverage planning, with the goal of minimizing travel time (Hameed et al., 2013). A variety of algorithms have been applied, including greedy algorithms (Oksanen & Visala, 2009), the Dragonfly algorithm (Muthukumaran & Sivaramakrishnan, 2019), genetic algorithms (Burchardt & salomon, 2006; Manikas et al., 2007), ant colony optimization (Akka & khaber, 2018), A* and dynamic A* (D*) (Guo et al., 2009). All these algorithms aim to determine the optimal route for navigation in the field- by using prior knowledge if it is available, or by preliminary mapping the area if there is no prior information on it, yet none suggest which plants should be sampled and in what order. due to limitations of resources such as time and energy, it is often impractical, because the robot still won't be able to sample all the plants field. In such scenarios, a sampling algorithm is necessary to prioritize and determine which plants should be sampled first.

# 3. Methods

The aim was to develop sampling algorithms that can be applied to a mobile robot monitoring an agricultural field to detect pests.

The proposed sampling algorithms were developed to take into account the insect spread patterns along with their spatial and temporal distributions and the field density. These algorithms were designed to save time by trying to reduce the number of sampling points by locating hot spots in the field. Two situations were characterized and examined in this research- when there is **prior information** available about the insects in the field (such as the type of insects, their spread rate, and dissemination patterns), and a situation where **no information** about the pests in the field is available. When there is prior information, the algorithm can be designed to utilize this information to make strategic decisions on sampling points and real-time adjustments to the sampling route. When there is no information the algorithm's ability to effectively detect pests in the field is impaired, as it is unable to make informed decisions on sampling points and real-time adjustments to the sampling route. In both situations, the goal is to locate as much insects as possible and minimize the number of sampling points (and thereby time), while ensuring accurate spatial and temporal sampling of the insects. The performance of these algorithms was evaluated for a multitude of conditions and compared to current state-of-the-art techniques.

## 3.1. Simulation environment

The simulation environment provided a controlled and repeatable environment for evaluating the performance of the proposed sampling algorithm. The simulated environment represents a large rectangular agricultural field with fixed shape, composed of rows of evenly spaced plants and with a consistent distance between each individual plant. The simulation environment for the sampling algorithms without prior information (chapter 4) were developed in MATLAB, while the simulation environment for sampling algorithms with prior information (chapter 5) were developed in Gazebo and RViz.

## 3.2. Sampling algorithms

The sampling algorithms aim to increase the amount of insects that the robot manages to detect. For the situation with no prior information (the location and distribution of insects is unknown), multiple algorithms were developed. For situations where prior knowledge of the insects in the field is available, a dynamic algorithm which was data-driven was developed to make decisions based on real-time data collected during the sampling process and compared to other state-of-the-art algorithms.

### 3.2.1. Sampling algorithm without prior information

Seven algorithms were developed to determine the robot's sampling path while traversing through an agricultural field to detect insects. Each algorithm represents a unique combination of strategy and method. The strategy refers to the manner in which the robot samples the plants in the field (from his right / left / both sides), and the method examines several ways to select the next plant for sampling, from a more thorough and local form to a form of expansion to larger areas in the field. The algorithms were evaluated for different scenarios and compared using multiple performance measures.

### 3.2.2. Sampling algorithm with prior information

The methods used in this research focused on the development and evaluation of a dynamic sampling algorithm for an agricultural monitoring ground robot, where prior information on the field and the insects in it, is available. The algorithm is designed to locate insects in a field where complete sampling of all the plants in the field is infeasible due to resource constraints. The dynamic algorithm is data-driven and utilizes real-time information to identify and prioritize sampling at suspected points, trying to locate hot spots and adapt sampling plans accordingly. The algorithm was implemented in a simulation environment, which was developed to evaluate its performance in several different scenarios, including extreme cases. The simulation settings were arranged to resemble an insect-infected field, and take into account a large number of input parameters, including the field size, the robot's speed, number of working days, number of viewpoints while sampling the plant, and more. In the algorithm evaluation stage, the dynamic algorithm was compared to two different algorithms, ("Naive" and "Bouncy", detailed below) to test its capabilities. The comparison was conducted according to three real scenarios derived from insect data collected in parallel research. Additional sensitivity analyses were conducted to investigate the effects of field size, insect spread rate, and detection time on the algorithm's performance. The performance of the algorithm was evaluated using various metrics including the number of hot spots detected, the time required to detect the hot spots, and overall accuracy.

### 3.3. Performance measures

The following performance measures were calculated, for each day and cumulatively until the end of the working days:

- Detection %- the percentage of insects located.
- Visited plant %- the percentage of plants sampled.

Additional performance measures were adjusted for each situation, as detailed in each chapter.

### 3.4. Analyses

For each situation multiple analyses were conducted. For the situation in which no a-prior information is available, sensitivity analyses were performed to focus on multiple parameters as follows:

- The size of field (0.2 ha, 0.4 ha, 0.8 ha)
- Disease severity: how fast the disease spreads (10%, 50%, 90%)

For the situation where prior information is available, three different scenarios were conducted to compare the three algorithms. The scenarios were formulated to focus on multiple parameters as follows:

- The size of the field (1 ha, 5 ha, 10 ha).
- Number of plants in the field.

The scenarios were based on data collected in previous research (Shmuel, 2018). The data included two-spotted spider mite (*Tetranychidae)* data based on a dataset acquired in Paran, Israel (Shmuel, 2018), which included monitoring data collected once a week from five different plots during the growing seasons of 2015–2017, with the pest's spreading behaviour tracked using GPS-GIS technologies. The simulation of the robot's traversal through the field, sampling the plants and trying to detect insects, was repeated 10 times for each scenario to ensure robustness and reliability of the results. The presented results are the average of all repetitions.

> Scenario A – Small plot size of 1 ha and 784 plants in the field.
> Scenario B – Medium plot size of 5 ha and 4096 plants in the field.
> Scenario C – Large plot size of 10 ha and 8281 plants in the field.

Sensitivity analysis was performed for the following parameters: plot sizes (from 1 to 20 ha), spread rates each day (30, 50 and 80%), detection times (20, 25, 30, 35 and 40 s) and robot detection rates (50, 60, 70, 80 and 90%).

# 4. Sampling algorithm without prior information

This chapter describes the development and evaluation of several sampling algorithms for a mobile ground monitoring robot in an agricultural field. The algorithms were first compared to reveal trends, followed by sensitivity analyses conducted to further evaluate the performance under different field sizes and disease spread severities. The results of these evaluations were used to make a comprehensive assessment of the proposed strategies in terms of their effectiveness, efficiency, and robustness.

## 4.1. Methods

A simulation was developed to test the efficiency of the seventh algorithms in various scenarios. The simulation was designed to simulate a field that has been infected by insects. The performance of the algorithms was compared across various scenarios and a sensitivity analysis was conducted to determine their performance.

## 4.2. Sampling algorithms

Seven distinct algorithms were developed. Each algorithm represents a unique combination of a sampling strategy and method.

The sampling strategy refers to the manner in which the robot samples the plants while advancing along the rows in the field: the "snake" strategy dictates that the robot will sample only one row of plants at a time, specifically the one to its right; in the "neighbor" strategy the robot samples the plants from two rows, those on its right and left.

The sampling method, describes how many plants the robot must forgo between samples. The methods studied were sampling every plant in the field, sampling every N plants in the field, and online random sampling where the robot skips a random number of plants (up to 4) at each step.

An additional algorithm was developed uses a distinct strategy - random sampling, whereby the robot will take samples of a fixed number of plants in the field in a completely random manner.

## 4.3. Simulation model

The simulation models were developed in MATLAB in a modular form for different sampling strategies and sampling methods within each strategy as detailed below. All code is detailed in Appendix A.

Snake strategies:

1. The robot samples every vine on it's right side.
2. The robot samples every n plants – only on it's right side.
3. The robot samples random plants (every 1-4 plants) – online random (where the number of plants sampled is random).

Neighbor strategies:

1. The robot samples every n plants –on it's right and left sides.
2. The robot samples random plants (every 1-4 plants) – online random (where the number of plants sampled is random).
3. The robot samples every vine –on it's right and left sides.

Random sampling:

4. The robot samples randomly 1/n of the field (the samples are randomly selected for each day a-priori)– random sampling.

### 4.3.1. Algorithm 1: Snake strategy, sample every vine

The robot samples every vine located on its right side. When the robot finishes one row- it turns to the next row on the right. The robot's movement path is presented in Figure 4 with the numbers indicating the sampling order.

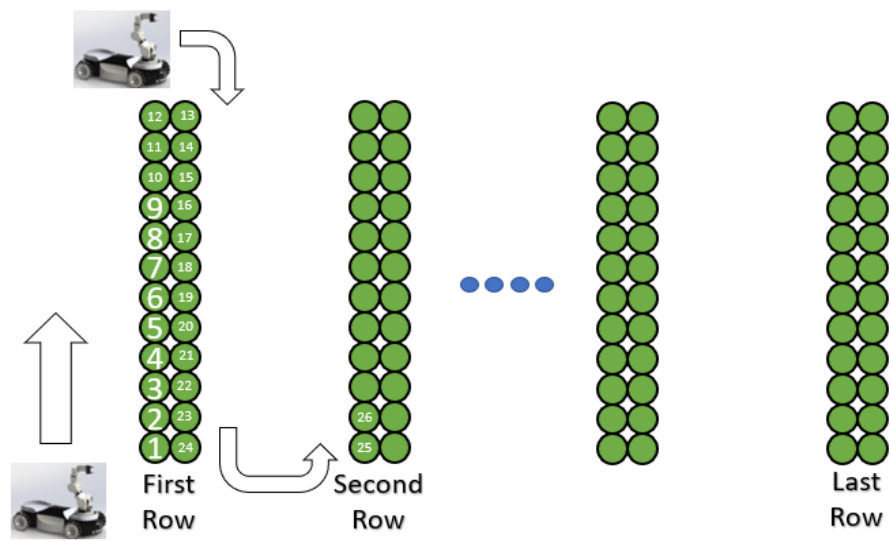

*Figure 3: Algorithm 1-Snake strategy, sample every vine*

### 4.3.2. Algorithm 2: Snake Strategy, sample every N plants

The robot samples every N plants, located on its right side. When the robot finishes one row- it turns to the next row on the right. When the robot finishes all the rows (or the last row), it turns around and retraces its steps from the first plant is did not yet visit. The robot's movements path is presented in Figure 5 with the numbers showing the sampling order for N=4.

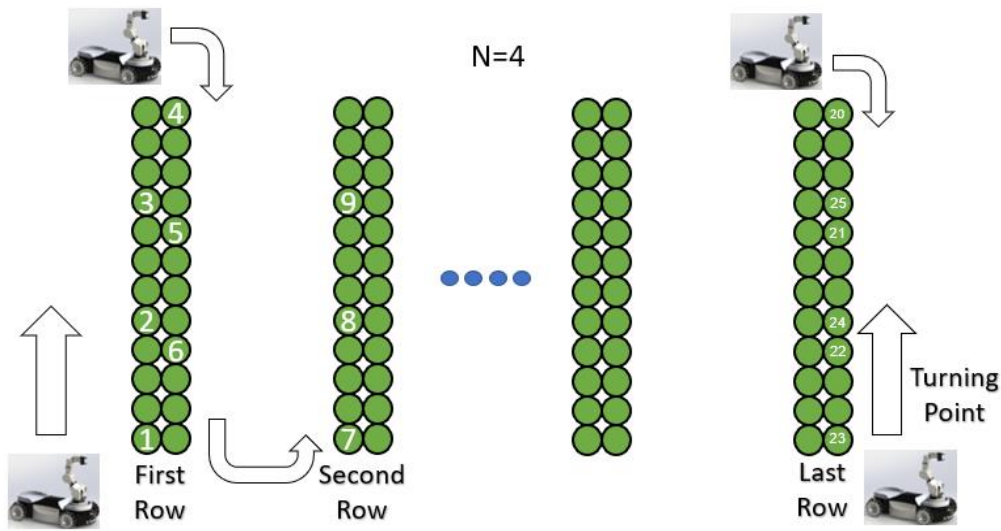

*Figure 4: Algorithm 2-Snake Strategy, sample every N plants – only on it's right side*

### 4.3.3. Algorithm 3: Snake Strategy, online random

The robot randomly samples plants, located on its right side. When the robot finishes one row- it turns to the next row on the right. When the robot finishes all the rows (or the last row), it turns around and retraces his steps from the first plant is has not yet visited. The robot's movements path is presented in Figure 6 with the numbers showing the order of the sampling.

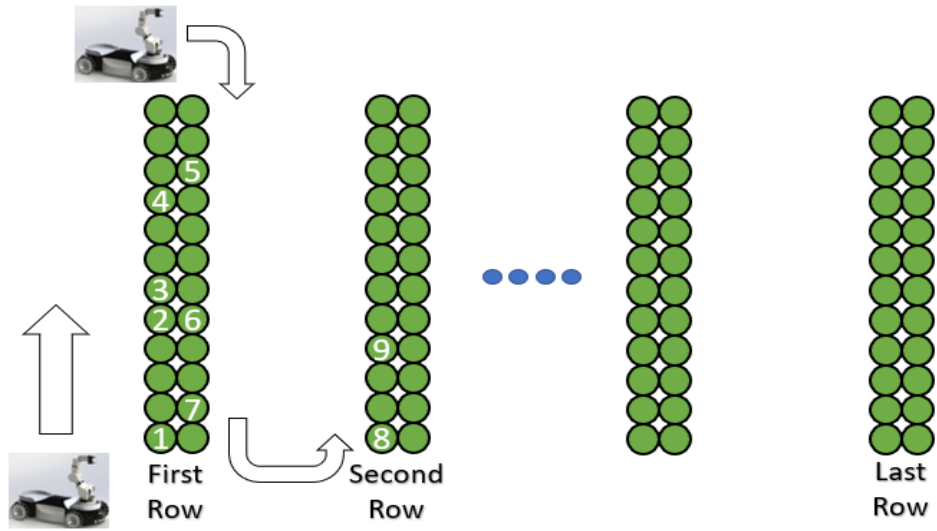

*Figure 5: Algorithm 3-Snake Strategy, online random, samples random plants*

### 4.3.4. Algorithm 4: Neighbor strategy, samples every N plants

The robot samples every N plants, located on its right side, and then samples its neighbor from the left side- except from the edges (first and last rows). When the robot finishes one row- it turns to the next row on the right. When the robot finishes all the rows (or the last row), it turns around and retraces its steps from the first plant it did not yet visited. The robot's movements path is presented in Figure 7 with the numbers showing the order of the sampling- in case N = 4.

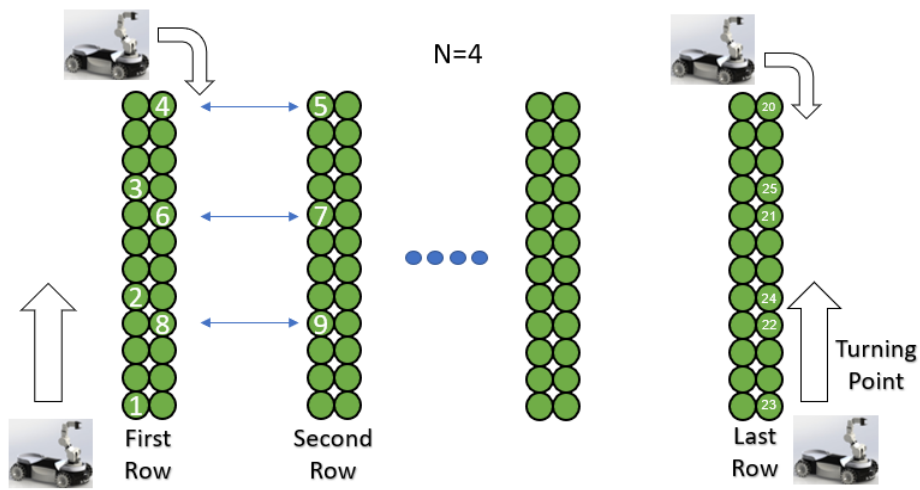

*Figure 6: Algorithm 4-Neighbor strategy, samples every N plants –on it's right and left sides*

### 4.3.5. Algorithm 5: Neighbor strategy, online random

The robot random plants, located on its right side, and then samples its neighbor from the left side- except from the edges (first and last rows). When the robot finishes one row- it turns to the next row on the right. When the robot finishes all the rows (or the last row), it turns around and retraces his steps from the first plant it did not yet visit. The robot's movements path is presented in Figure 8 with the numbers show the numbers shows the order of the sampling.

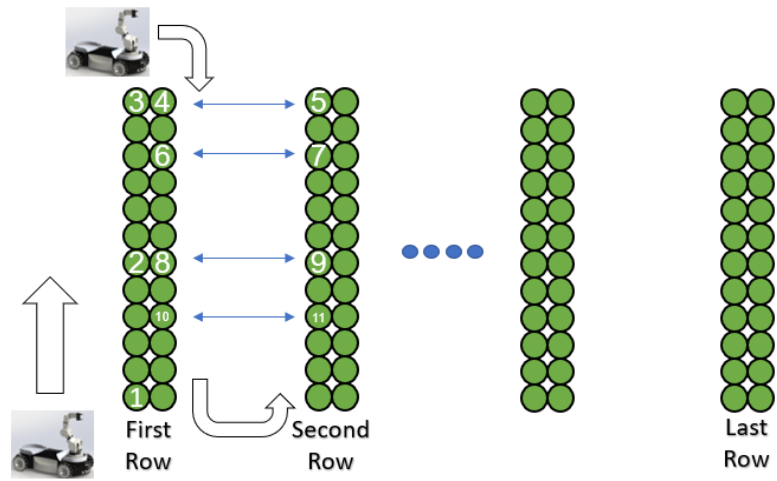

*Figure 7: Algorithm 5-Neighbor strategy, online random, samples random plants*

### 4.3.6. Algorithm 6: Neighbor strategy, sample every vine

The robot samples every vine located on its right side. When the robot finishes one row- it turns to the next row on the right. The robot's movement path is presented in Figure 9 with the numbers indicating the sampling order.

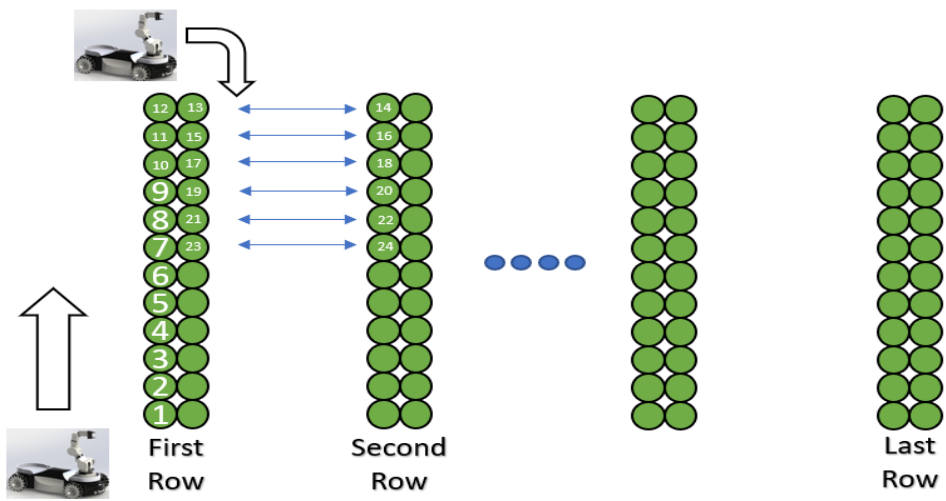

*Figure 8: Algorithm 6-Neighbor strategy, sample every vine*

### 4.3.7. Algorithm 7: Random strategy

The robot samples randomly 1/N of the field (the samples are randomly selected for each day a-priori)– random sampling. The robot's movement path is presented in Figure 10, and relevant only for day 1. At day 2 the sampling will be completely random again.

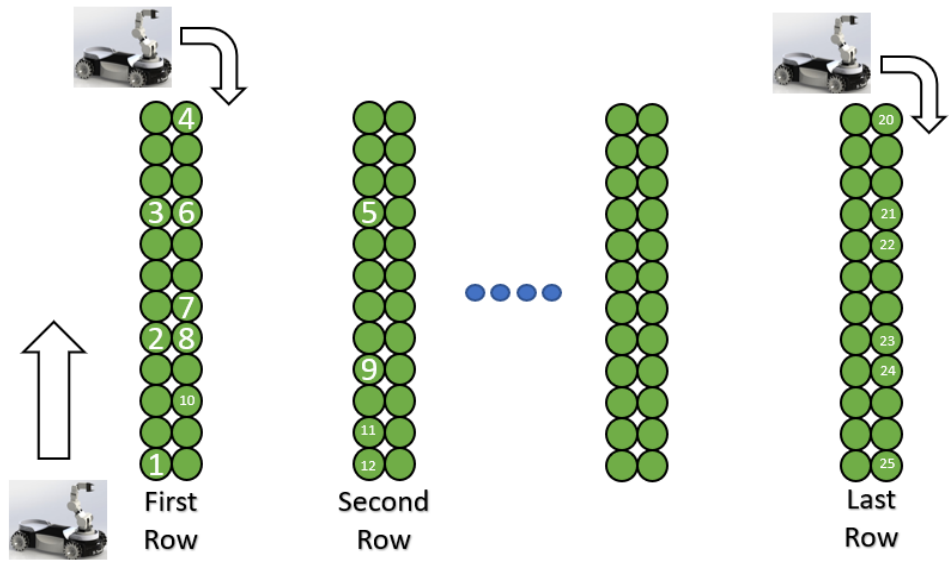

*Figure 9: Algorithm 7- Random strategy*

### 4.4. Simulation environment

The robot must travel between rows without harming plants, and from row to row as fast as possible. Turning around at end of row is conducted automatically. The common vineyard structure (Figure 11) was assumed to include an area of about 2 dunams. The growing areas consist of several rows of seedlings, in each row there are two adjacent seedlings. The distance between the rows is 3.5 meters, and the distance between neighboring seedlings is 2 meters. These values were selected based on a commercial vineyard in Lachish, central Israel.

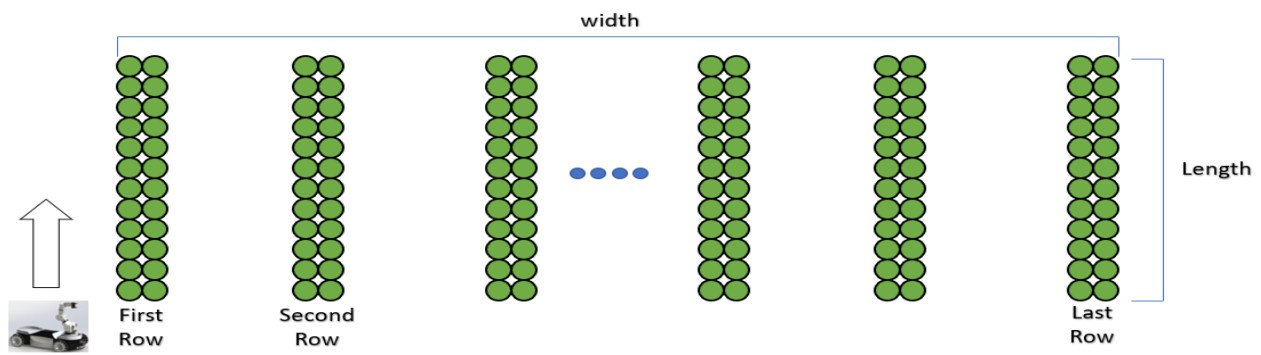

*Figure 10: Simulation environment*

### 4.4.1. Assumptions

- Each grower has one robot.
- The robot travels through the field while sampling the plants for insects.
- Every row of plants has two sides, except for the side rows (first and last).
- At the end of each day the insects spread to nearby plants; spread is according to severity level.
- If a plant with insects is sampled, it is assumed the insects are detected.
- If insects are detected, they are sprayed and destroyed (and do not spread further).
- At the end of every day, the robot will stop its work and continue from the same point the following day.
- Dimensions and speeds: initial values were assumed as detailed below.
- When the robot reaches the end of the field- it turns around and starts sampling using the same algorithm from the other side of the vineyard until the day is completed (and this is repeated if needed from the other side).

### 4.4.2. Performance measures

The following performance measures were defined:
- Percentage of Visited Plants each day – PVV_ED
- Percentage of Correct Detections each day - PCD_ED
- Percentage of Visited Plants all working days – PVV_ALL
- Percentage of Correct Detections all working days - PCD_ALL
- Missed detections - MD
- Time to reach 30%, 50%, 80%, 100% detection – how many days does it take to detect 30%, 50%, 80%, 100% of current disease spread – D30, D50, D80, D100

### 4.4.3. Variables

- LOR = Length of rows
- WOR = Width of rows
- Days = number of working days
- DistPlants = Distance Between Plants (meters)
- Dunam = Dunams the farmer has. 1 Hectare = 0.1 dunam => 1 dunam = 1000m^2
- Robot_Speed_KMH = Robot Speed - Kilometers per hour
- Robot_Speed = Robot speed – (Meters per Second)
- DiseaseProbability = Chance of having a bug on the vine
- DiseaseSeverity = Defines how fast distributes to neighbor plants- the spread next day
- NumOfPlantsInRow = Number of plants in a single row
- NumOfRows = Number of rows in the field
- NumOfRowsToCheck = Rows to check- each row has two lines of plants- except of the edges

- ❖ oneVPTime = Total seconds for one Viewpoint
- ❖ twoVPsTime = Total seconds for two Viewpoints
- ❖ threeVPsTime = Total seconds for three Viewpoints
- ❖ turningTime90 = Total seconds for turning to the plants/back to road 90 degrees
- ❖ turningTime180 = Total seconds for turning 180 degrees to the neighbor
- ❖ totalTime = Total second robot works during the day
- ❖ timeLeft = Total seconds left for the rest of the day
- ❖ timeBetweenPlants = Navigating time between two adjacent plants- seconds
- ❖ timeBetweeenRows = Navigating time between two adjacent rows- seconds
- ❖ numOfDetections = Count total number of Correct Detections
- ❖ numOfDetectionsToday = Count number of Correct Detections each day
- ❖ TotalBugsToday = Count number of bugs at the beginning of a day

### 4.4.4. Steps in performing the simulation

The simulation was developed in a modular way to enable to obtain results depending on the field characteristics selected by the user at the beginning of the simulation.

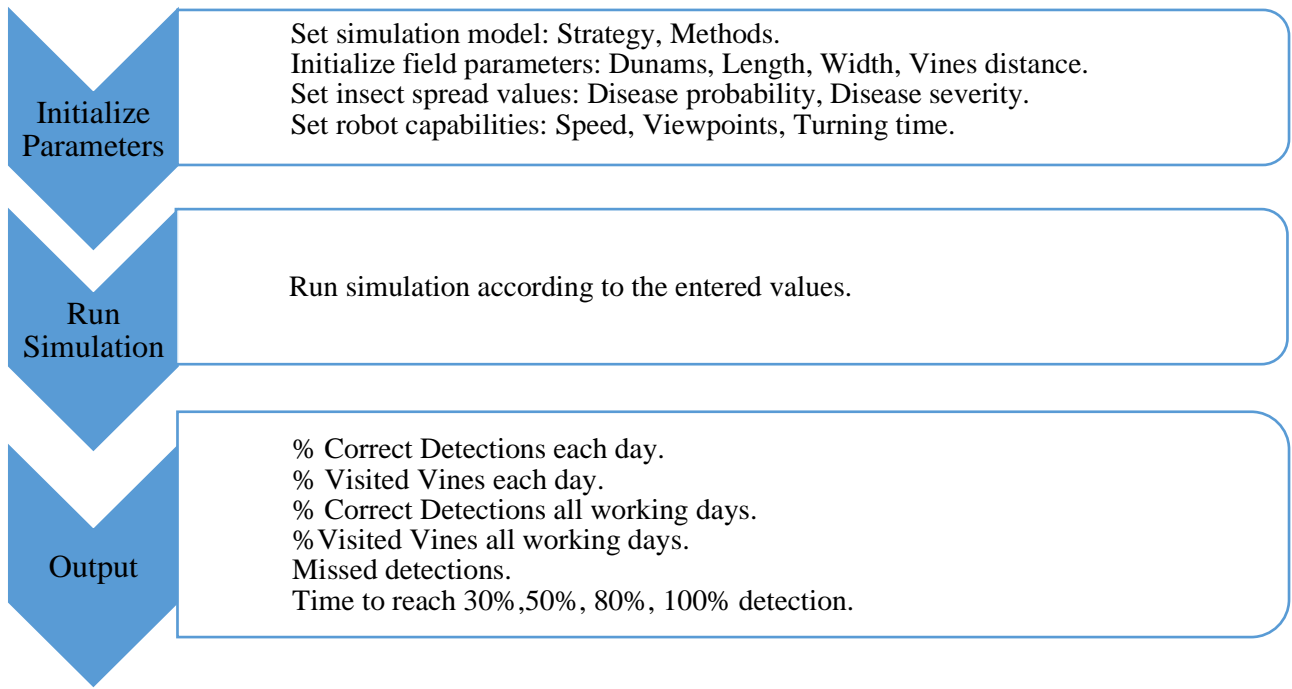

**Initialize Parameters**
Set simulation model: Strategy, Methods.
Initialize field parameters: Dunams, Length, Width, Vines distance.
Set insect spread values: Disease probability, Disease severity.
Set robot capabilities: Speed, Viewpoints, Turning time.

**Run Simulation**
Run simulation according to the entered values.

**Output**
% Correct Detections each day.
% Visited Vines each day.
% Correct Detections all working days.
%Visited Vines all working days.
Missed detections.
Time to reach 30%,50%, 80%, 100% detection.

## 4.5. Analysis

### 4.5.1. Basic analysis

The constant values selected for the initial simulation were based on data collected from an expert grower in Lachish farms:

- ❖ LOR = 100 (Length of rows)
- ❖ WOR = 3.5 (Width of rows)
- ❖ days = 3 (number of working days)
- ❖ DistPlants = 2 (Distance Between Plants)
- ❖ Dunam = 2 (this was evaluated for 4 and 8 also)
- ❖ robot speed km\h = 2.5 KM\H (Travel speed along row - includes accelerations; reaches maximum speed after 0.75 meters)
- ❖ DiseaseProbability = 0.3
- ❖ DiseaseSeverity = 0.3 (this was evaluated for 0.1, 0.5 and 0.9 also)
- ❖ VP time:
- ❖ oneVPTime = 4.34 (Total seconds for one View Point)
- ❖ twoVPsTime = 16.14 (Total seconds for two View Points)
- ❖ threeVPsTime = 23.74 (Total seconds for three View Points)

- ❖ turningTime90 = 10 (Total seconds for turning to the plants/back to road 90®)
- ❖ turningTime180 = 23 (Total seconds for turning 180® to the neighbor)
- ❖ totalTime = 7,200 (total seconds robot works during the day- 2 hours)
- ❖ timeBetweeenRows = 25 seconds (time between rows)

### 4.5.2. Sensitivity analysis

Sensitivity analyses were performed for size of field and disease severity as noted below.

- ❖ Size of field: basic size (Scenario A) vs. x2 growth (Scenario B) and x4 growth (Scenarios C).
- ❖ Disease severity: how fast the disease spreads 10%, 50%, 90% (corresponding to scenarios D, E, F).

## 4.6. Results and discussions

### 4.6.1. Basic analyses

All the results presented refer to the basic inputs (Length of rows = 100 m, Width of rows = 3.5 m, Distance between plants = 2 m, Dunam = 2, Disease probability = 0.3) unless otherwise noted, except for severity, which defines the spread level.

First, the percentage of detection of infected plants for each day and each algorithm was tested for two severity levels (low – 0.1, high-0.9). Results for low level of spread (Figure 12) reveal that only 5 algorithms reach 100% of detection at the end of 3 working days. Four of these five algorithms are algorithms that work in the Neighbor strategy (except for the online random algorithm), and the fifth algorithm is the algorithm which works according to the snake method (which samples each vine). In addition, there are two algorithms that achieve 100% of detection after 2 working days- the neighbor algorithm that samples every vine (algorithm 6), and the neighbor algorithm which samples every second vine (algorithm 4). Furthermore, the online random algorithm that executes the neighbor strategy, shows better results from day 1 till the end of the working days, from every algorithm with the snake strategy (except of algorithm 1, that samples every vine).

At the high level of spread, there are only three algorithms that reach 100% detection after 3 working days (Figure 13)- algorithm 6 (neighbor strategy, samples every vine), algorithm 4 (neighbor strategy, samples every second vine) and algorithm 1 (snake strategy, samples every vine). Similar to the low level, they reach 100% of detection even at the end of the second day. Although they all yield perfect detection accuracy at the end of the second day, it can be seen that at the end of the first day, the algorithms that use the neighbor strategy reach 20% more accuracy than the algorithm that uses the snake strategy.

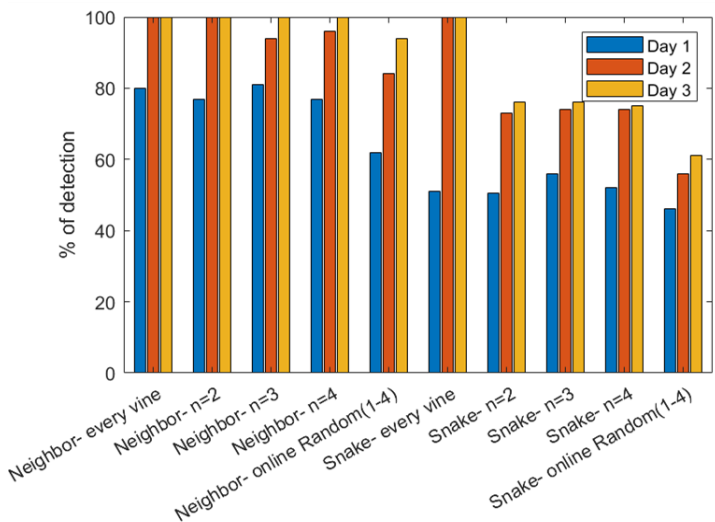

Figure 11: Algorithms comparison- percent detection, low spread level (10%)

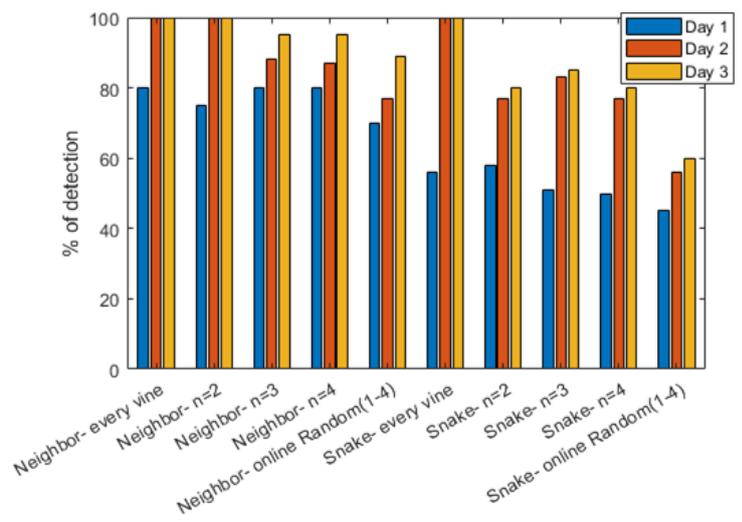

Figure 12: Algorithms comparison- percent detection, high spread level (90%)

Figures 14 and 15, show analysis of the number of days to reach specific detection percent (D30, D50, D80, D100) for different algorithms. The difference between the figures is the severity level of the spread, which is low in Figure 14, and high in Figure 15. At the high level of spread, we can see that most of the snake algorithms yield more than 50% of detection after the first working day, but because of the high rate of spread, they do not reach 80% even after the third day. Unlike them, the results of the first two neighbor's algorithms (samples every vine and every second vine), do not change and yield 100% at the second day.

The remaining three algorithms in this strategy find it difficult after the first working to get to the perfect accuracy but do present better results after the first day than the snake strategy.

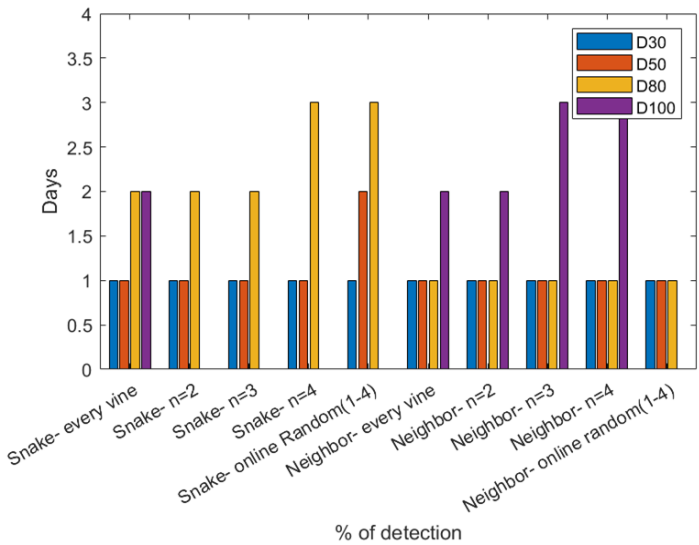

*Figure 14: Algorithms comparison- days to reach specific percent detection, low spread level (10%)*

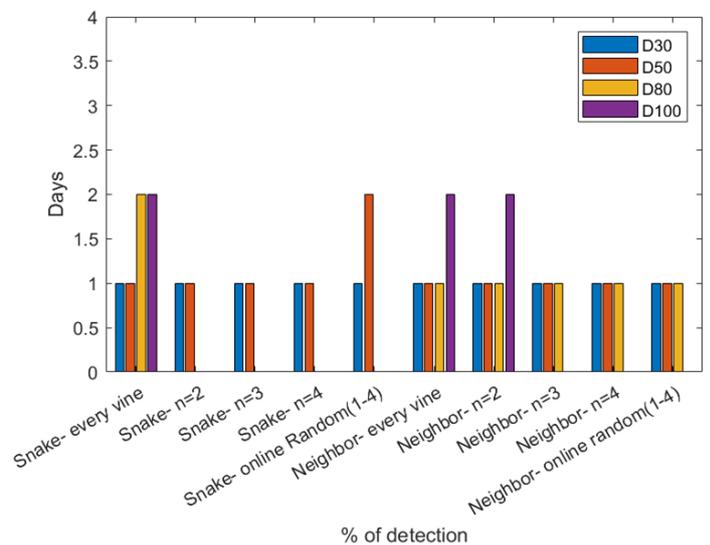

*Figure 14: Algorithms comparison- days to reach specific percent detection, high spread level (90%)*

In order to compare the random algorithms, which samples exactly 1/N plants every day (implying they sample one field after n days), larger fields were tested (random 1/4 and online random (1-4): LOR = 200, Dunam = 6; Random 1/8 and online random (1-8): LOR = 210, Dunam = 11). The figures below show the results for two severity levels- Figure 16 for low level (0.1) level and Figure 17 for high level (0.9). Comparing the results between the two figures reveal similar results showing that the algorithm that yields best detection percentage is algorithm 5 (Neighbor strategy, online random). On the other hand, at high rate of disease spread, among the algorithms that was tested on the larger field (8) there was no significant difference, and there is no preferred algorithm.

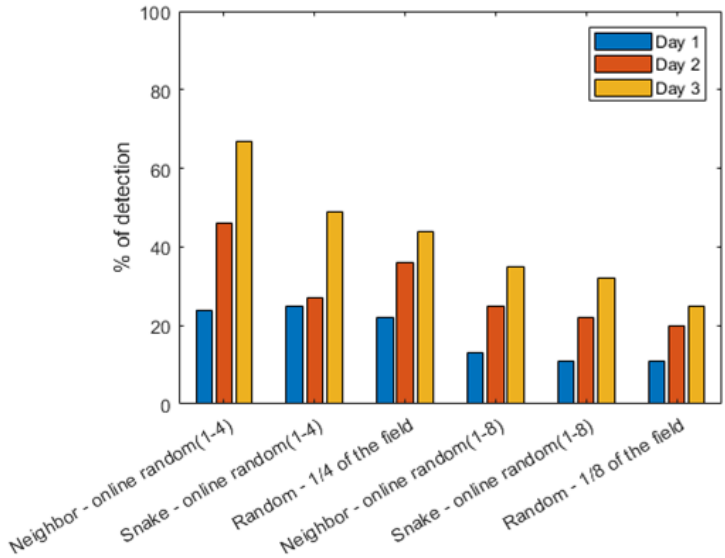

*Figure 16: Comparison algorithms with random (1/4- 1/8), low spread level (10%)*

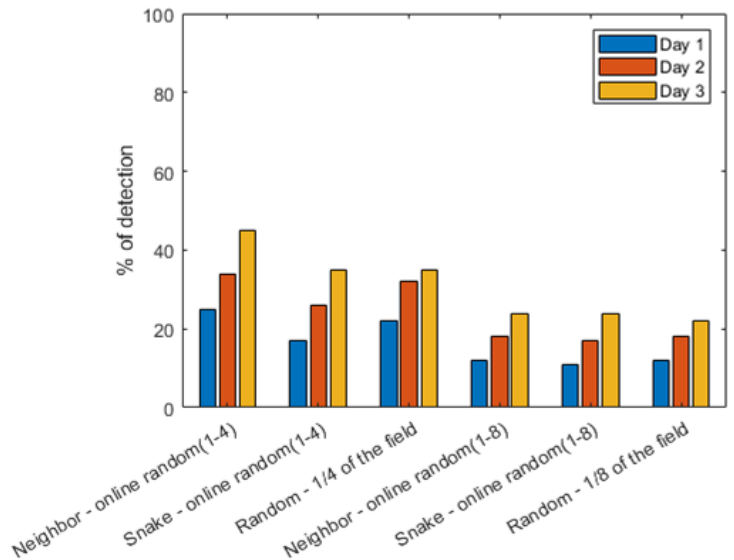

*Figure 15: Comparison algorithms with random (1/4- 1/8), high spread level (90%)*

In order to compare the random algorithms, which samples exactly 1/n plants every day (implying they sample one field after n days), with the neighbor and snake strategies, larger fields were tested (Random 1/4: LOR = 200, Dunam = 6; Random 1/2: LOR = 200, Dunam = 4). As seen in the figures below, which checks different severity rates- Figure 18 for low level (0.1) and Figure 19 for high level (0.9), the only algorithm that yields 100% of detection, is the neighbor algorithm that samples every second vine. As the number of plants between two different samples increases, and the field size increases, the success capabilities will decrease accordingly.

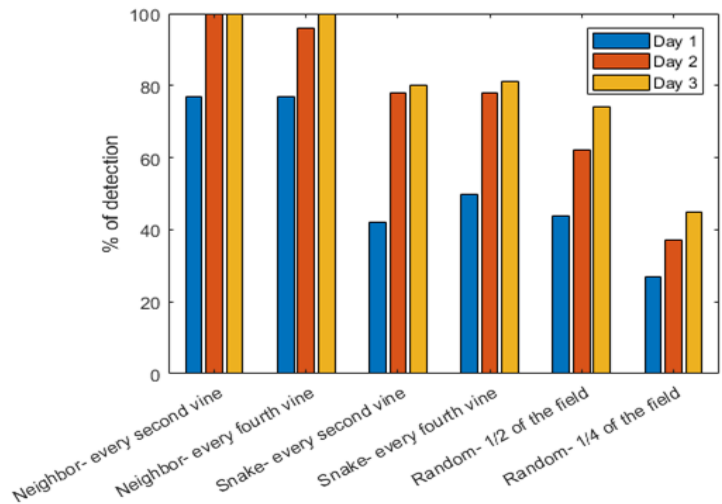

*Figure 17: Comparison algorithms with random (1/2- 1/4), low spread level (10%)*

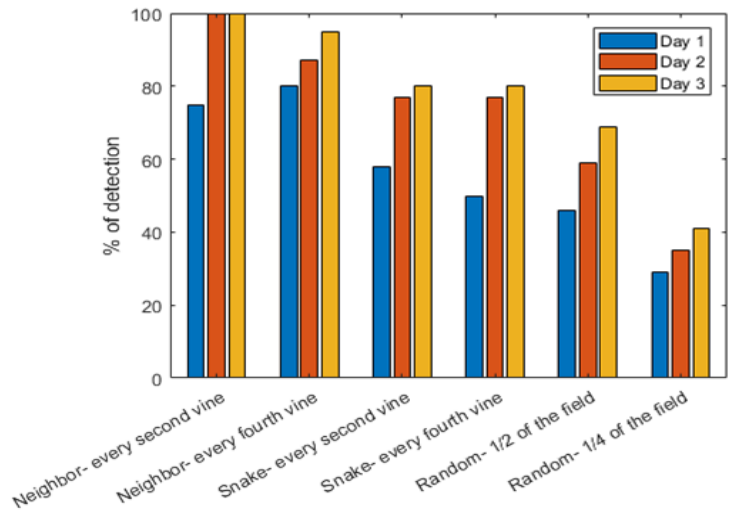

*Figure 19: Comparison algorithms with random (1/2- 1/4), high spread level (90%)*

### 4.6.2. Trend analysis

The percentage of detection of infected plants for each day and each algorithm are presented for the two strategies evaluated for the most common scenario, scenario A (simulation input):

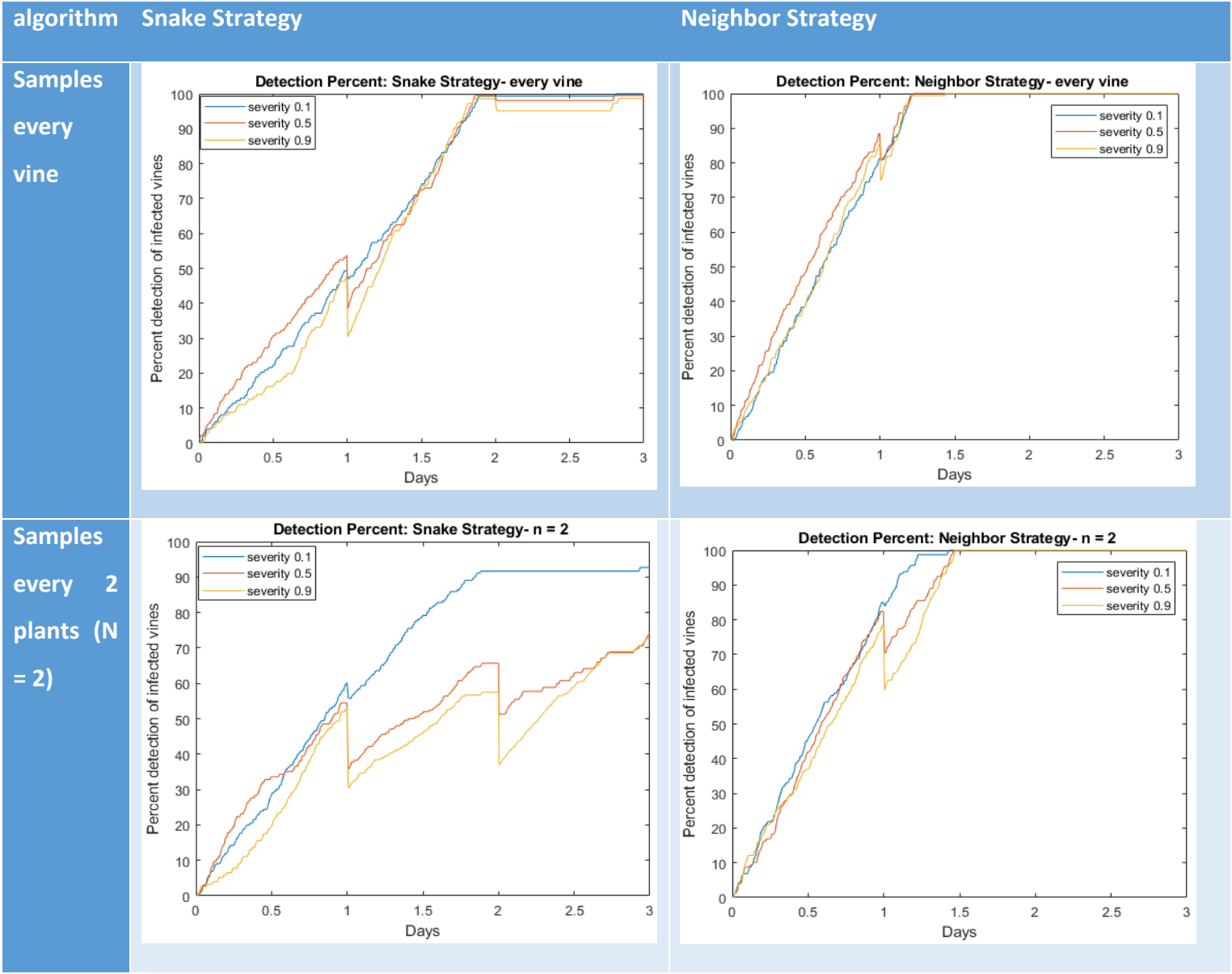

| algorithm | Snake Strategy | Neighbor Strategy |
|---|---|---|
| Samples every vine | | |
| Samples every 2 plants (N = 2) | | |
| Samples every 3 plants (N = 3) | 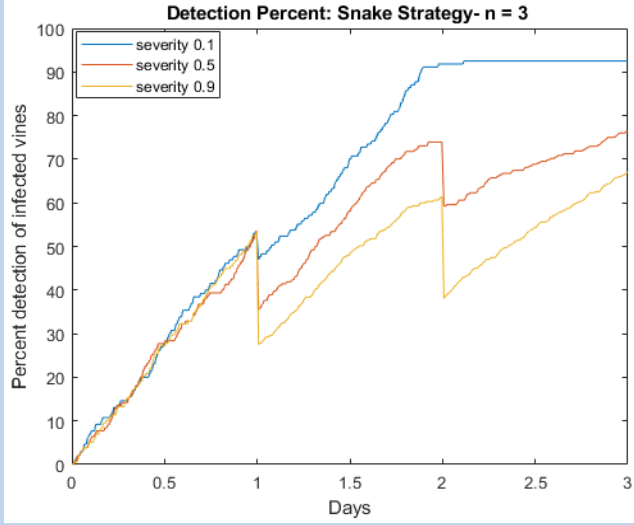 | 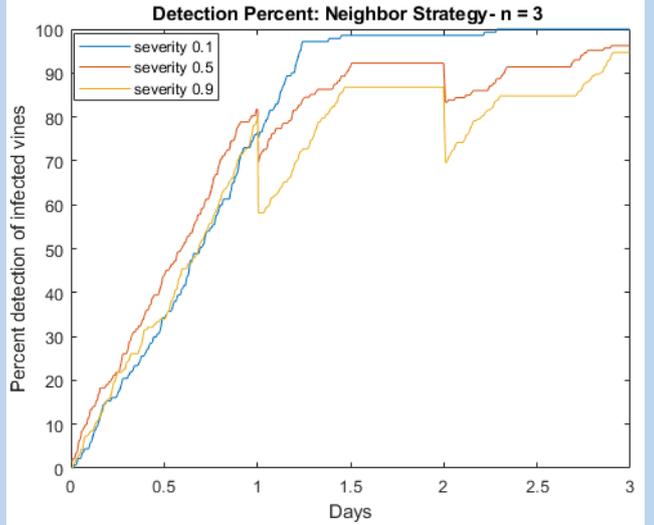 |
| Samples every 4 plants (N = 4) | 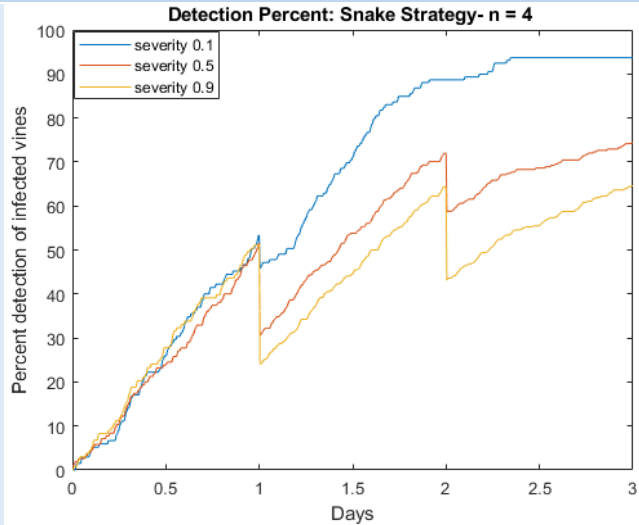 | 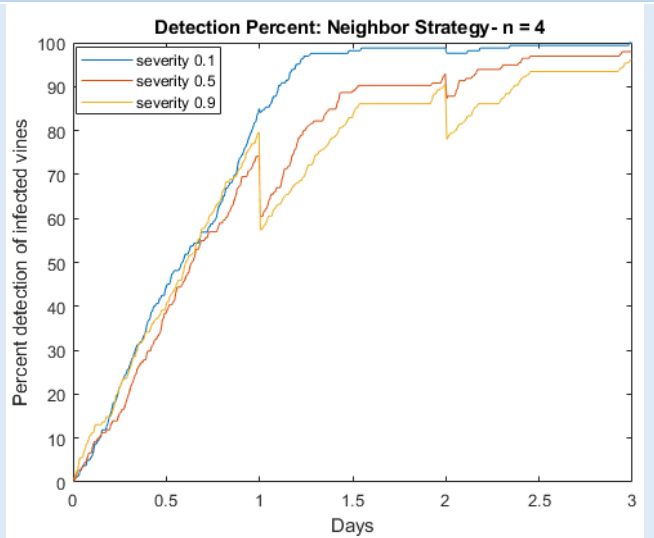 |
| Samples random plants | 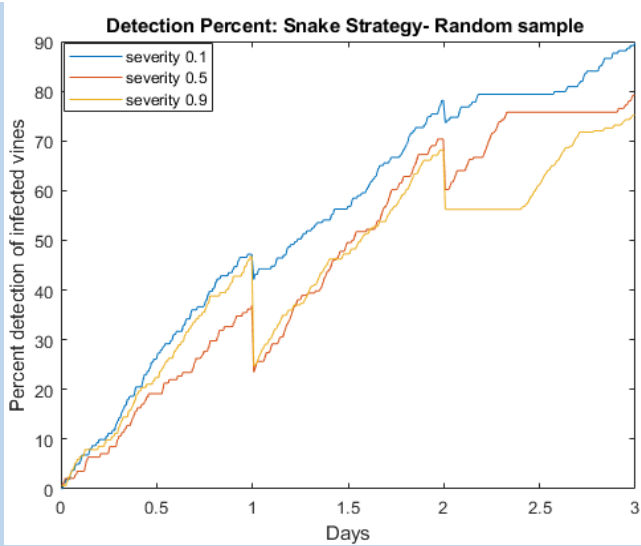 | 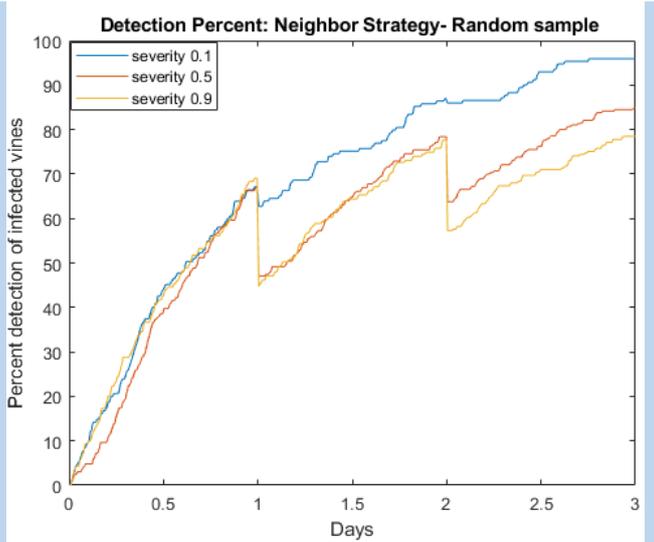 |

*Table 1: Trend Analysis- algorithms comparison*

The trend analyses reveal:

- The fastest algorithm that reaches 100% detection is algorithm 6- which executes the neighbor strategy and does not skip any vine- samples every pair of neighbors plants. It yields 100% after 2 working days.
- After each working day, there is a drop in the detection percentages, depending on the spread severity. This fall is due to the spread of insects at the end of each day (depends on the severity rate). Due to the form of spread of the insects, which spreads to the plants adjacent to them, we will conclude that the more insects there are at the end of the day - the greater the spread. Therefore, in most of the graphs in which the neighbor strategy was triggered, the fall was smaller, implying fewer insects spread. This is because each day the robot manages to detect a larger amount of insects as compared to the snake strategy (Appendix A, Table 2).
- Comparing the two strategies (Snake/Neighbor), neighbor strategy achieves 20-40% higher detection rate each day, which indicates of a larger slope in the neighbor strategy, as compared to the snake strategy implying a higher detection rate in this strategy.
- The neighbor strategy does not appear to have substantial differences in the final results for the different disease severity rates (0.1, 0.3, 0.5, 0.9), where the std stands for 6% at the most. Unlike the snake strategy which is effective for small severity rates but shows a relatively low percentage of detection for higher spread chances with std that can reach the 30%.
- In all the algorithms except for the random algorithm, there are curves where the slope is equal to 0 for a long working time, which indicates that the robot makes a turn and repeat it's footsteps (at the beginning or end of the row), thus it visits the same plants once again, indicating inefficiency. Those flat curves are significantly larger in the snake strategy over the neighbor strategy.

Additional results (Appendix A, Tables 2-4) reveal:

- Algorithms in which an online random visit is conducted (algorithms 3,5) do not reach the best and expected result for two measures:
    - Percentage of Visited Plants all working days- the algorithms present lower than expected results, around 85% in each algorithm, implying that after 3 days the robot did not visit all the plants.
    - Percentage of Correct Detections all working days – As a direct result of the previous section, the robot did not find all the insects at the end of the working days and find only 75% of the insects.
- Random algorithms, which visit a fixed number of plants (half of the field/ quarter of the field etc..), and samples them randomly, shows worse results even than when compared to online random algorithms, due to their lack of systematicity.
- The only algorithms that enabled to visit all plants and detect all infected plants, and after three working days were not left with any insects in the field are algorithms 1 (Snake Strategy- samples every vine), 4 (Neighbor strategy, samples every 2 plants) and 6 (Neighbor strategy, samples every vine).
- Probably as a result of the simulated insect spread, the best algorithms in these categories, are the algorithms in which the plants are tested without skipping any vine- or at the most check every 2 plants, thus preventing the miss of one insect on the way, which causes spread the next day to nearby plants.
- Comparing these three algorithms by the other performance measures- Percentage of visited Plants each day and Percentage of Correct Detections each day shows interesting results. Although the snake strategy (algorithm 1) presents worse results regarding the number of visited plants every day compared to the neighbor strategy (algorithms 4,6), it shows better percentage of correct detections every day- around 50, than these algorithms, which presents approximately 40% of correct detections.
- In the large field (scenario C, field of 80 dunam):
    - There is only one algorithm that reaches 100% success- algorithm 6 (Neighbor strategy, samples every vine), this is achieved after 3 days.
    - The snake strategy yields 50% detection success only at the 3$^{rd}$ working day, and does not reach 80%, in any algorithm.

### 4.6.3. Conclusions

For regular fields of 2 dunam size, with only 1 working robot, it is not necessary to visit a large quantity of plants every day, but the sampling must be done thoroughly, and plants should not be skipped or at least as little as possible. Algorithms that visit every vine, or at the most every 2 plants, yield the best detection rates, over algorithms that skip plants (samples every third/fourth vine). Accordingly, and as expected the higher the number of plants between samples- the lower the success rates.

In larger fields- 4-8 dunams, we will probably prefer algorithm 6, the algorithm that executes the neighbor strategy and samples every vine, which performs thorough work, and at the same time controls a larger amount of plants on each working day (around 60% every day, compared to 25% in the snake strategy). This algorithm has the highest chance of reaching 100% detection, no matter how large the field is.

In case there are constraints, and the sampling cannot be on every vine (n > 1), the best strategy is the neighbor strategy. Although in the two strategies the robot fails to reach perfect identification after three working days, it can be clearly seen that the success rates of the robot in the neighbor strategy (algorithm 4) are significantly higher (90%) than the snake strategy (algorithm 2 – 60%).

# 5. Sampling algorithm with prior information

The algorithm is data-driven, utilizing real-time information to identify and prioritize sampling at suspected points, trying to locate hot spots and adapt sampling plans accordingly. A simulation environment was constructed to examine the algorithm's performance under various growth and insect spread conditions, and it was compared to two existing algorithms. To simulate actual data we used *Tetranychidae* insect location data from a previous research. Additional sensitivity analyses were conducted to investigate the effects of field size, insect spread rate, and detection time on the algorithm's performance.

## 5.1. Methods

A simulation environment was developed to evaluate the performance of the dynamic algorithm in several different scenarios, including extreme cases. The simulation settings were arranged to resemble an insect-infected field. In order to test the capabilities of the dynamic algorithm it was compared to two different algorithms. The comparison was conducted according to three scenarios derived from insect data collected in parallel research. Furthermore, a sensitivity analysis was carried out in order to evaluate how the algorithm would perform in different situations.

## 5.2. Sampling algorithms

The proposed dynamic sampling algorithm was compared to two common algorithms, 'Naive' and 'Bouncy'.

### 5.2.1. Proposed algorithm- dynamic sampling

The leading principle of the dynamic sampling algorithm (Fig. 18) is to focus mainly on the areas where it is more likely to find an insect; if the detection result is positive at a specific location, more weight is given to sampling in nearby plants in an attempt to locate large areas of spread (hot spots), in an efficient and thorough manner.  Initially, two lists are created: 'Open list' for the plants that are planned to be visited, where from every point the robot's next sample will be the closest plant in the list, and 'Close list' for the plants that have already been visited. A plant that is inserted into the 'Open list' is one that has been identified as suspicious based on real-time data, and prior information such as insect-spread pattern. As soon as an insect is detected, the robot adds nearby plants to the 'Open list'. The sampling loop is performed until the end of the workday, or until the 'Open list' is empty. When the list becomes empty, unvisited suspicious plants are added to the list.

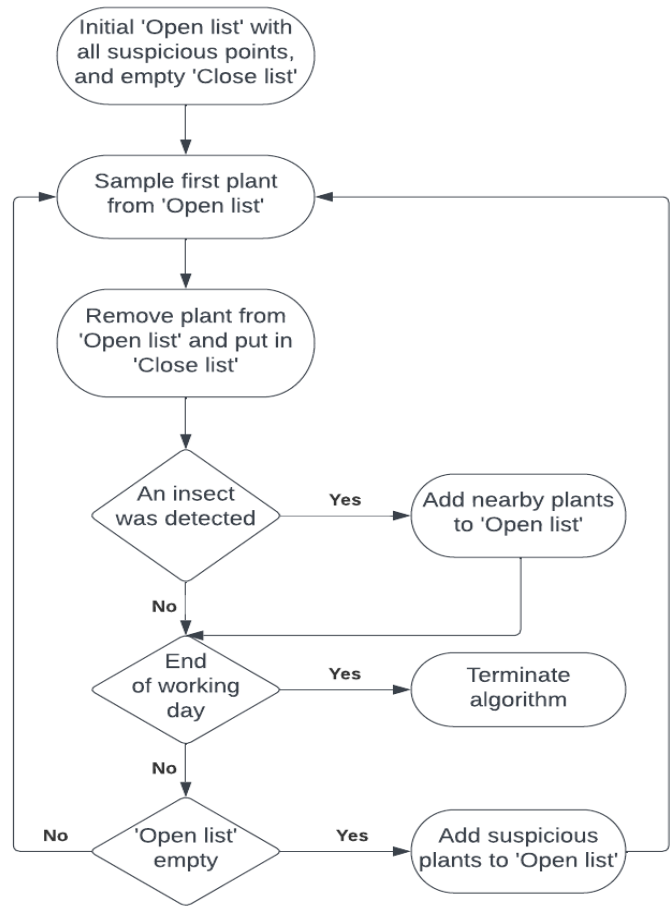

Figure 18: Dynamic Sampling Algorithm flow chart

### 5.2.2. Naive and Bouncy algorithms

In the 'Naive' algorithm, the robot samples every plant in the field, one by one, in a predetermined order. In the 'Bouncy' algorithm, the robot samples every N plants in the field in a predetermined order (for example, when N = 2, half of the plants in the field will be sampled). Throughout this article, the 'Bouncy' algorithm was used with an N value of 2.

### 5.3. Simulation environment

A virtual environment was developed in the Gazebo simulator. The robot's motion is visualised in RViz software, to give a better understanding of the environment, without any need to work in the field or to use real insects. In addition, this enables transferring the algorithms to an actual robot system as-is as soon as it is developed. Fig. 19 presents the robot in the Gazebo and RViz environments.

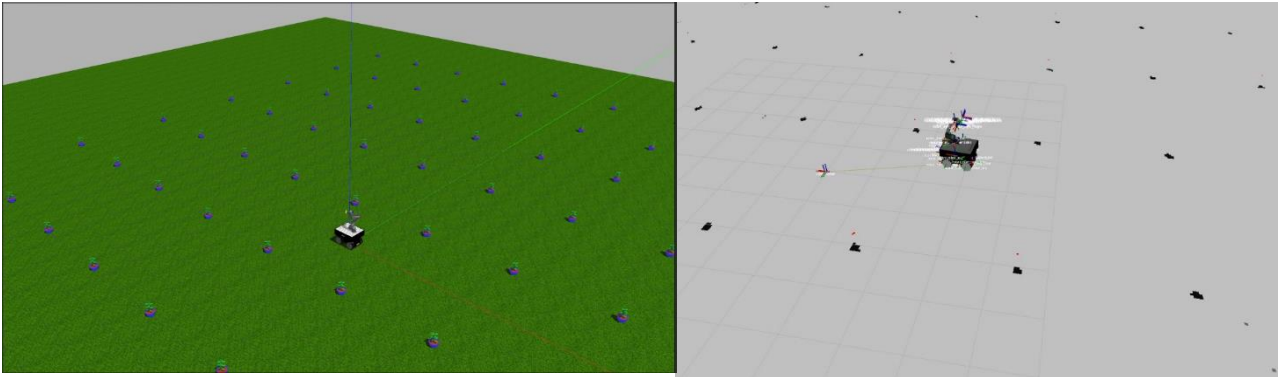

*Figure 19: Simulation environment- Gazebo (left) and RViz (right)*

### 5.3.1. Assumptions

The following assumptions were considered when developing the simulation model.
- Each grower has only one robot, for a defined duration each day (working day).
- The paths are free of obstacles.
- The robot travels along each row and samples the plants.
- Whenever a row ends, the robot moves forward to the next row.
- The robot cannot cross rows and must drive to the end of each row to move on.
- To perform the inspection, the robot stops next to the plant that is being sampled.
- When the robot reaches a plant, it inspects the plant from several viewpoints in order to detect the insects; the more viewpoints sampled, the longer the robot stays next to the plant. The addition of inspection viewpoints increases the inspection detection performance (reliability).
- If an insect is detected, it is destroyed and does not spread any further.
- Plants can be sampled only once each day, but multiple times during the simulation (along different days).
- If the insect is not detected, it can spread to adjacent plants, whether along the same row or across to the next row, at the end of the day.
- The robot might miss an insect during inspection.

### 5.3.2. Parameters

The following independent parameters were used in the simulation as default parameters.
- The robot works 25 days, 12 h per day.
- The robot's average speed is 2.5 km/h.
- The distance between adjacent rows is 4 m with a 3 m distance between plants.
- Values for sampling times and detection rates were linked as detailed below:
  - When the time spent searching for an insect is less than 20 s (less than one viewpoint), the detection rate will be 0%.

- When the time spent searching for an insect is 20 s (one viewpoint), the detection rate will be 50%.
- When the time spent searching for an insect is 25 s (two viewpoints), the detection rate will be 60%.
- When the time spent searching for an insect is 30 s (three viewpoints), the detection rate will be 69%.
- When the time spent searching for an insect is 35 s (four viewpoints), the detection rate will be 77%.
- When the time spent searching for an insect is 40 s (five viewpoints), the detection rate will be 84%.
- When the time spent searching for an insect is more than 40 s (more than five viewpoints), the detection rate will be 90%.
- The robot spends 40 s next to each plant that it samples, and the detection rate is 84%.

## 5.4. Analysis

### 5.4.1. Scenarios

Three different scenarios were conducted to compare the three algorithms. The scenarios were formulated to focus on multiple parameters as follows:

- The size of the field (1 ha, 5 ha, 10 ha).
- Number of plants in the field.

The scenarios were based on data collected in previous research [33]. Two-spotted spider mite (*Tetranychidae*) data were based on a dataset acquired in Paran, Israel [33], which was based on monitoring data collected once a week from five different plots during the growing seasons of 2015–2017, and the pest's spreading behaviour was tracked using GPS-GIS technologies [34]. The simulation of the robot's traversal through the field, sampling the plants and trying to detect insects, was repeated 10 times for each scenario to ensure robustness and reliability of the results. The presented results are the average of all repetitions.

Scenario A – Small plot size of 1 ha and 784 plants in the field.

Scenario B – Medium plot size of 5 ha and 4096 plants in the field.

Scenario C – Large plot size of 10 ha and 8281 plants in the field.

### 5.4.2. Sensitivity Analysis

Sensitivity analysis was performed for the following parameters: plot sizes (from 1 to 20 ha), spread rates each day (30, 50 and 80%), detection times (20, 25, 30, 35 and 40 s) and robot detection rates (50, 60, 70, 80 and 90%).

### 5.4.3. Performance measures

The following performance measures were calculated for each day and cumulatively until the end of the working days:

- Percent detection – the percentage of insects located.
- Percent plants visited – the percentage of plants sampled.

## 5.5. Results and discussion

### 5.5.1. Algorithms comparison

#### 5.5.1.1. Scenario A

For Scenario A, all three algorithms reached 100% detection at the end of the working days, with the goal being reached after 2 days for both the dynamic sampling and 'Naive' algorithms and after 4 days for the 'Bouncy' algorithm (Fig. 20).

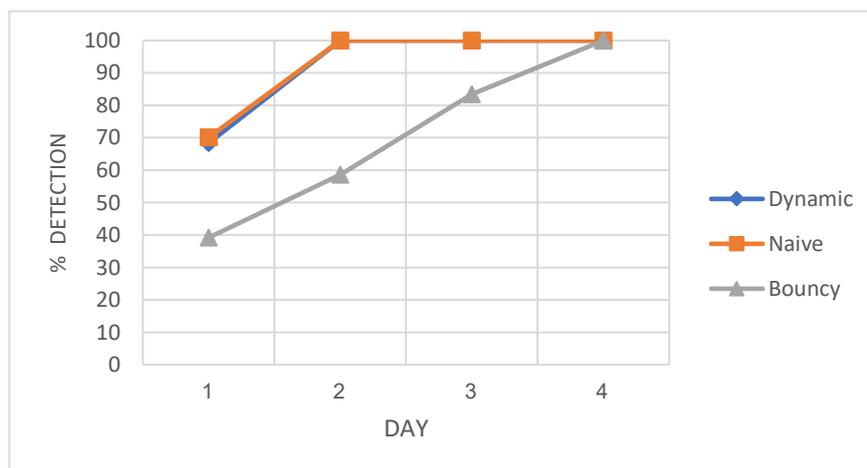

*Figure 20: Algorithms comparison- Percent detection – Scenario A.*

Fig 21. presents the percentage of plants out of all of the plants in the field that were sampled by the robot, each day. As expected, the more plants the robot sampled on each working day, the faster the algorithm reached 100% detection. The 'Naïve' algorithm, which sampled the largest number of plants each working day (an average of 67% of the plants), and the dynamic sampling algorithm (an average of 53%) achieved 100% detection after 2 days of work, whereas for the 'Bouncy' algorithm (an average of 47% of the plants each day), it took 4 days to discover all of the insects in the field.

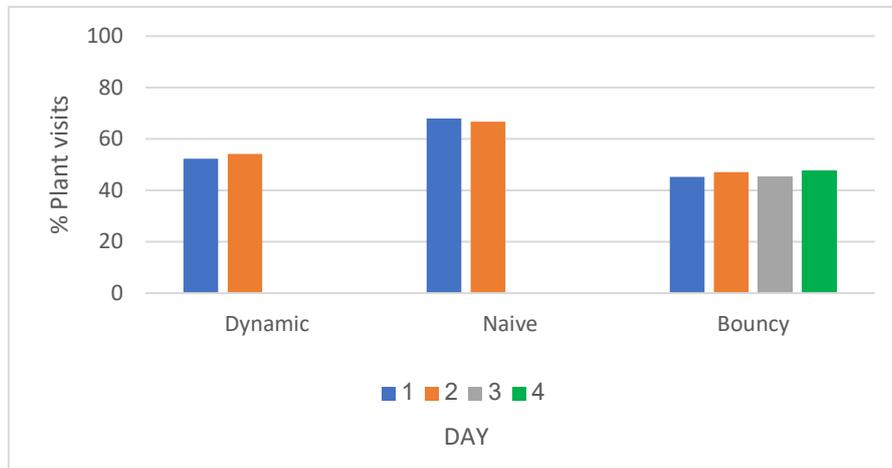

*Figure 21: Algorithms comparison- Percent plants visited – Scenario A.*

### 5.5.1.2. Scenario B

For Scenario B, there was a significant difference between the algorithms in percent detection, with the dynamic sampling algorithm achieving 84% detection at the end of the working days, compared to 68% for the 'Naïve' algorithm and 46% for the 'Bouncy' algorithm (Fig. 22).

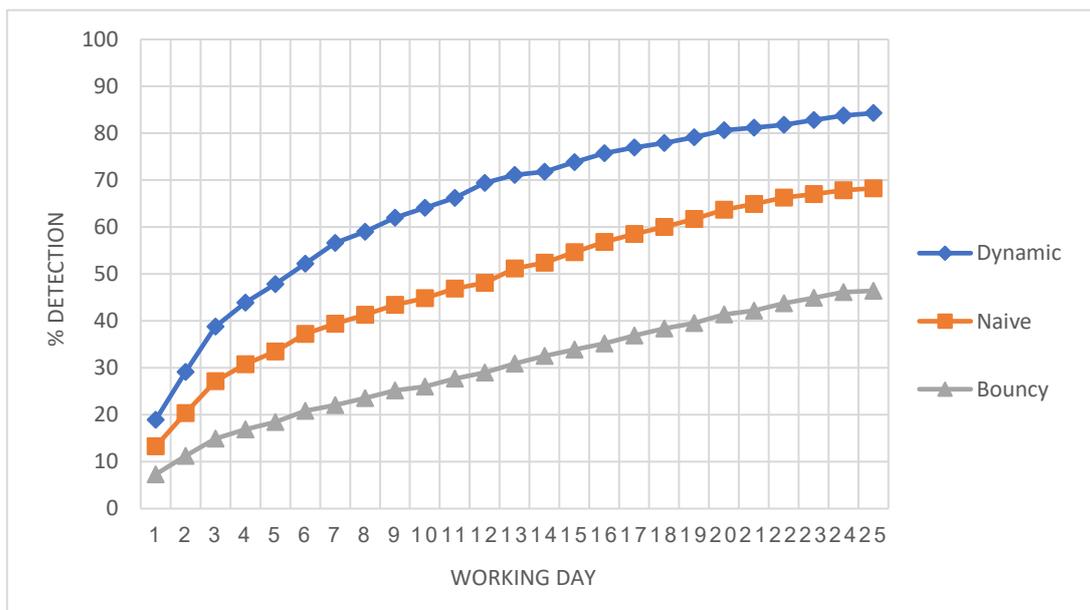

*Figure 22: Algorithms comparison- Percent detection – Scenario B.*

Analysis of the percentage of plants out of all of the plants in the field that were sampled by the robot, each day (Fig. 23) revealed an interesting insight: it is not necessary to sample many plants every day to ensure detection of a large number of insects. Although the dynamic sampling algorithm achieved the most rapid and best results, it was not the one that sampled the largest number of plants each day. On the other hand, while the 'Naive' algorithm succeeded in sampling a larger number of plants during each working day separately (average of 15%), it gave worse results than the dynamic sampling algorithm (average of 12%). The shorter the distance travelled by the robot between samples, the less time it spends travelling in the fields and the more time it devotes to sampling, thus managing to reach a larger number of plants. Therefore, the number of visits with the dynamic sampling algorithm on the first days was more similar to that with the 'Naive' algorithm. For every sample in which an insect is detected with the dynamic sampling algorithm, the robot will not skip plants, but as the days pass, and the number of insects in a field decreases, the robot will skip plants in order to locate hot spots and will behave more like the 'Bouncy' algorithm.

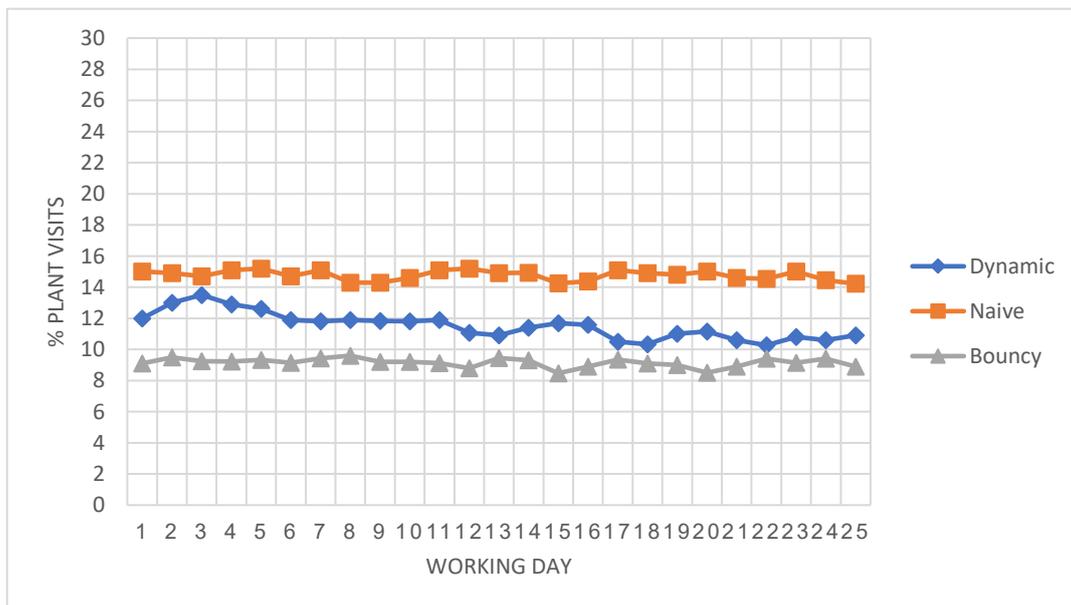

*Figure 23: Algorithms comparison- Plant visits % - Scenario B.*

### 5.5.1.3. Scenario C

Results for Scenario C revealed a significant difference between the three algorithms in percent detection (Fig. 24). The 'Bouncy' algorithm gave the worst results (maximum 22% detection) due to the low number of plants sampled each day, even though on the first days it showed very close performance to the 'Naive' algorithm. In contrast, the 'Naive' algorithm showed better results and consistent improvement over time, but at the end of the working days reached only 34% detection. The dynamic sampling algorithm achieved the best results with 49% detection at the end of the working days.

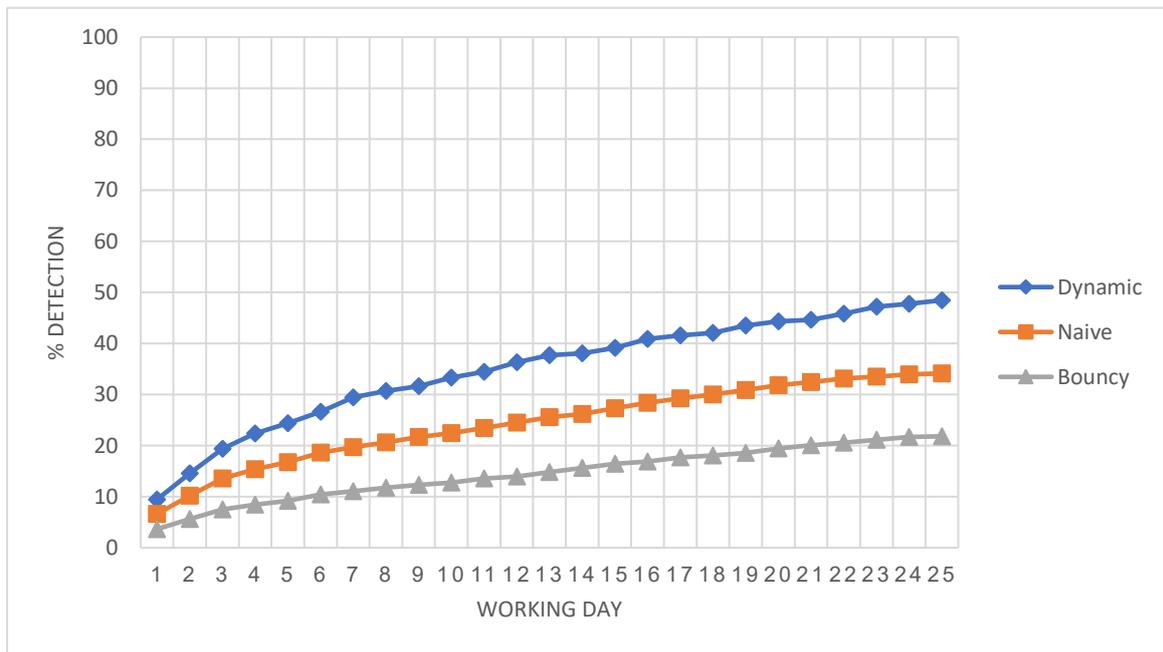

*Figure 24: Algorithms comparison- Percent detection – Scenario C.*

## 5.5.2. Sensitivity analysis

For the default parameter values, the dynamic sampling algorithm already produced excellent detection results of 100% after about a week in a 1 ha field (Fig. 25) For most regular-sized fields (2–5 ha), it reached up to 90% detection. In large-size fields, the robot failed to detect the same number of insects as in the smaller fields, and as the field size increased, there was a moderate decrease in the percentage of detection at the end of the working days.

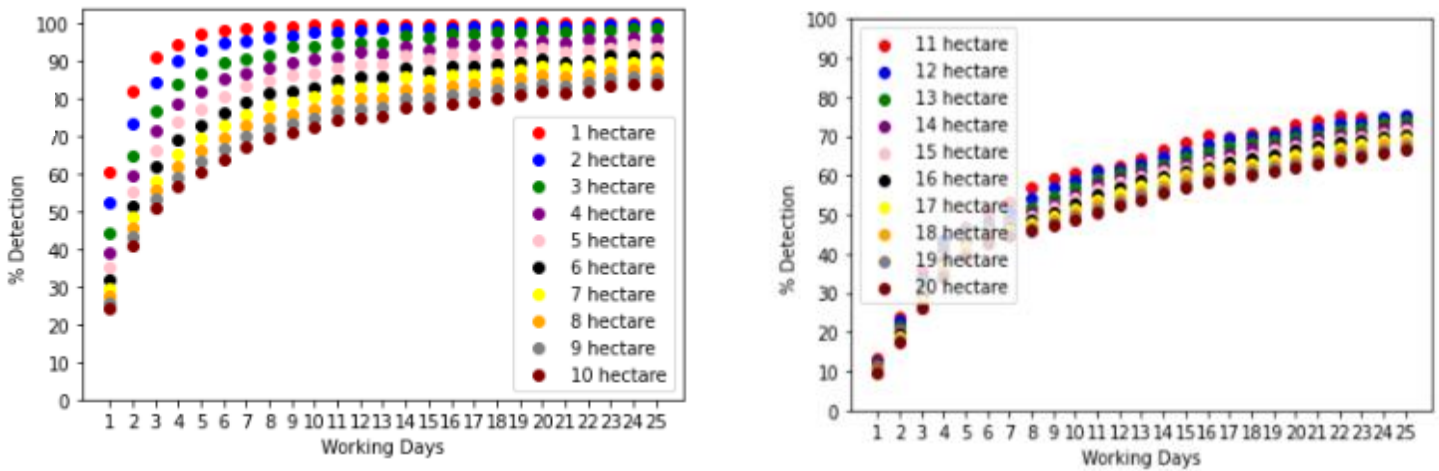

*Figure 25: Dynamic algorithm analysis- Percent detection in fields of different sizes.*

As expected, as the size of the field increased, the detection percentage decreased (Fig. 26). The robot achieved a high detection result (90–100%) in small-sized fields, but the percentages decreased sharply in fields between 7 ha and 14 ha. In larger fields, detection percentage was characterised by a moderate linear decrease.

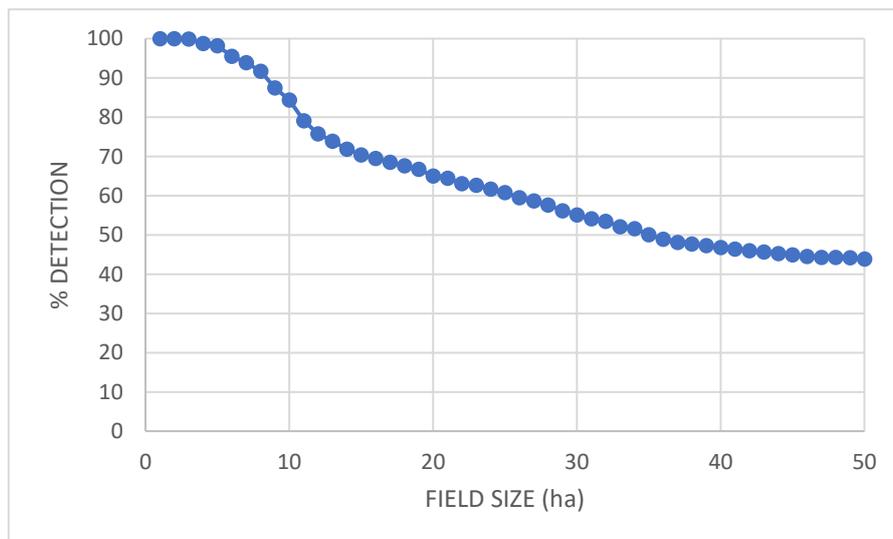

*Figure 26: Dynamic algorithm analysis- Percent detection vs. field size.*

The insects' rate of spread did not influence the final detection rates (Fig. 27). Although on the first few days there was a minor difference between the three rates of insect spread, where a lower rate of spread produced an average of 5% more detection, the results of the robot were the same at the end of the working days for each of the different field sizes. This stems from the way in which the algorithm works, i.e., locating hot spots in the field and acting accordingly.

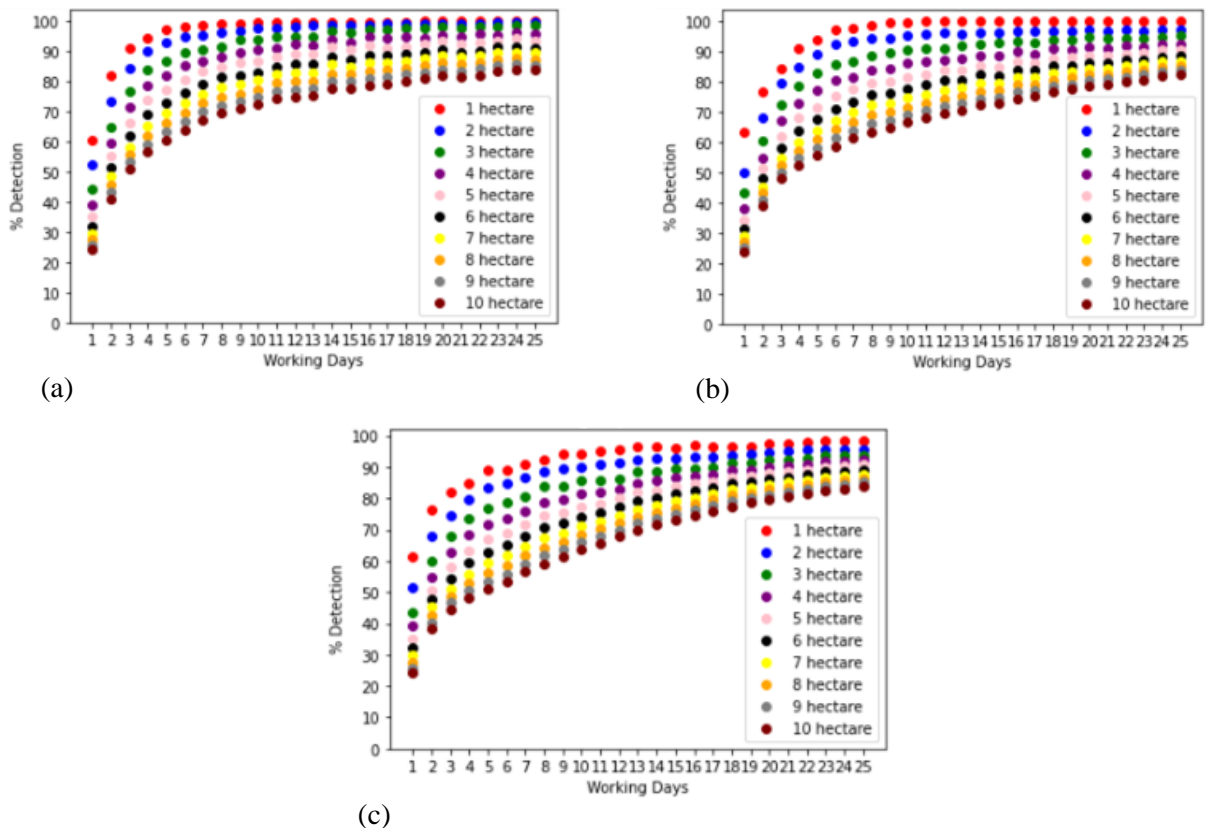

*Figure 27: Dynamic algorithm analysis- Percent detection for different spread rates. (a) 30% (b) 50% (c) 80%.*

If an insect exists on each plant, the more time the robot invests searching for it, the greater the chance that it will find it. The trade-off between time spent searching for insects in each plant and the detection percentage was also tested (Fig. 29). Results revealed that for a field of 6 ha, even if the number of plants that are inspected each day is small, the more time the robot spends searching for insects in each plant, the more insects will be detected. Thus, already on the first day, the robot managed to detect 50% more insects if it doubled the time spent on each plant (32% in 40 s versus 21% in 20 s). On the other hand, with time, the differences became smaller, and at the end of the working days, all cases reached over 90% detection.

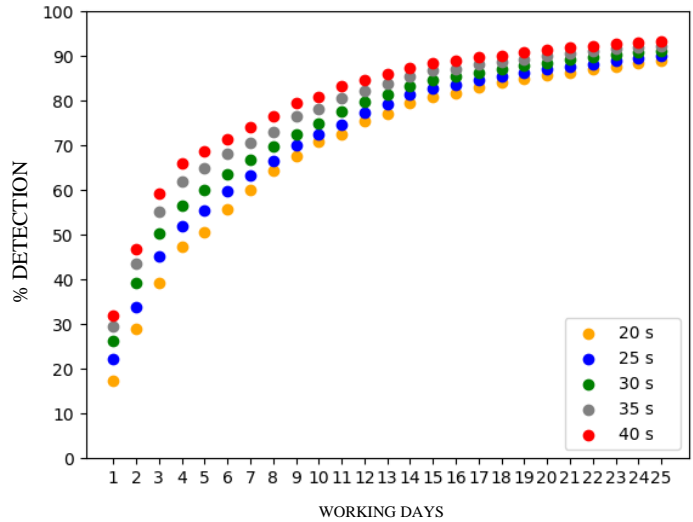

*Figure 29: Dynamic algorithm analysis- Percent detection for different visit times.*

Fig. 28 presents the results for robots with different detection rates, where different robots (or the same one, with different detection capabilities) use the dynamic sampling algorithm. As expected, better detection capabilities led to better detection results. However, the dynamic sampling algorithm allowed up to 75% detection, even for robots with a very low percentage of detection ability (40%).

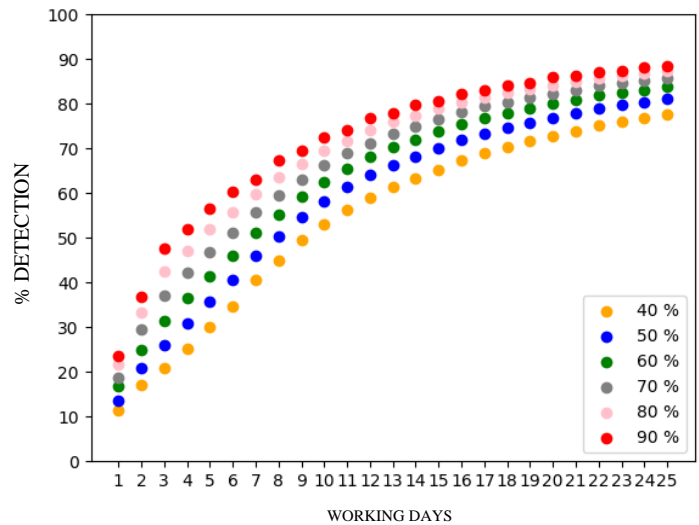

*Figure 28: Percent detection –for robots with different detection rates*

## 5.6. Conclusions

A dynamic sampling algorithm for monitoring insects in agricultural fields under limited resources—time, money or energy, among others—is proposed to determine an efficient way for a mobile ground robot to sample an insect-infested field. The dynamic sampling algorithm uses real-time data from the field and decides where to sample based on a-priori information and on the plants that it samples along the way. By applying this approach, it is possible to increase detection rates, regardless of field size and robot detection capability.

The comparison of the dynamic sampling algorithm to two other existing algorithms, on real data collected in previous studies, revealed its better performance. Although for a small field of 1 ha, its result was similar to that of the 'Naive' algorithm, in more complex cases of larger fields or higher rates of insect spread, it presented up to 50% better results for the detection of insects in the field.

In the sensitivity analyses, the dynamic sampling algorithm was tested against different rates of insect spread (simulating different insects) and presented identical results, regardless of the spread rate. The detection capabilities of the robot were also tested. Although on the early working days, it was evident that the better the robot's detection capabilities, the more effective the algorithm would be, over time, it was proven that the robot's capability was not an influencing factor; at the end of the simulation, the algorithm showed similar results, regardless of how good the robot's insect-detecting capability was. These comparisons would have been impossible without the use of simulation.

We therefore recommend utilisation of the proposed dynamic sampling algorithm in agricultural monitoring applications as it demonstrates great performance in terms of detection rate and time efficiency when compared to existing algorithms, and its results are robust to variations in field size, insect spread rate and detection times. Real-world implementation on an actual mobile robotic system is underway.

# 6. Conclusions and future work

This research developed and compared sampling algorithms for an insect monitoring robot travelling along an agricultural field. In this thesis, the problem was divided into two main situations: when prior information about the field and the insects is not available and when there is a-priori information on the field.

When there is no prior information, several important observations were made. For small fields, sized at 2 dunams with only one robot, thorough sampling is crucial, and skipping plants should be minimized. Algorithms that visit every vine or every second plant result in the best detection rates compared to those that skip plants. As expected, the higher the number of plants between samples, the lower the success rate. For larger fields, ranging from 4-8 dunams, the neighbor strategy which samples every vine was found to be the preferred approach as it performs thorough work while controlling a larger amount of plants each day (about 60% compared to 25% in the snake strategy). This algorithm has the highest chance of reaching 100% detection, regardless of the field size. In scenarios where sampling every vine is not feasible due to constraints, the neighbor strategy was found to be the best alternative. Although both strategies fail to reach perfect identification after three working days, it is clear that the success rate of the neighbor strategy (90%) is significantly higher than that of the snake strategy (60%).

The developed dynamic algorithm designed for fields with prior knowledge on insect populations, presenting significant findings. This algorithm utilizes real-time data from the field and makes decisions on sample collection based on the plants it has already sampled. This approach can lead to an increase in detection rates, regardless of the size of the field or the detection rate. The comparison of the dynamic sampling algorithm to two existing algorithms on real data from previous studies showed its superior performance. Although for a small field of 1 hectare it produced results comparable to the 'Naive' algorithm, in larger fields or with higher rates of insect spread, the dynamic algorithm demonstrated a higher rate of insect detection in the field, reaching up to 50%. Sensitivity analyses were performed to test the dynamic sampling algorithm against different rates of insect spread and the results were consistent, regardless of the spread rate. The detection capabilities of the robot were also evaluated, and although initially it appeared that the better the robot's detection capabilities, the more effective the algorithm would be, it was ultimately determined that this was not a determining factor. Regardless of the quality of the insect detection algorithm, the results of the simulation at the end showed similar outcomes. Additionally, the dynamic sampling algorithm was evaluated against different types of detection algorithms with varying detection capabilities. At the beginning of the simulation, there was a substantial difference in the detection rates (around 30%), but by the end, the difference was reduced to 10%, resulting in a 75% detection rate even with the worst detection algorithm (with a 40% detection rate).

This study introduces preliminary steps toward implementing a sampling algorithm in a mobile robot to monitor insect populations in agricultural fields. However, there are some limitations that must be addressed in future work. For instance, the study does not consider the crucial aspect of coordinating a fleet of robots working together in the field. Additionally, the study is focused on only one specific insect, the Two-spotted spider mite, and further evaluations are needed to determine the algorithm's effectiveness on a wider range of insects, pests, diseases, and viruses. These limitations highlight the need for further research to fully realize the potential of using mobile robots and sampling algorithms for monitoring insect populations in agriculture.

Future work in this research could include several avenues of exploration. Firstly, it would be valuable to examine insect spread according to a vector spread of each insect, which is influenced by various environmental factors such as weather, wind, insect type, and time of day. By characterizing insect spread in this manner, a more nuanced understanding of the behavior of insects in the field could be developed. Additionally, it may be worthwhile to investigate the performance of the algorithm when analyzed on an hourly rather than a daily basis. This would allow for a more fine-grained analysis of the algorithm's effectiveness, particularly in instances where insects are more active during certain times of the day. Furthermore, the creation of a map of insect probability in each plant, based on previous sampling, could provide valuable insights into the distribution of insects in the field. This would enable more targeted sampling efforts and could potentially improve detection rates. Future studies should determine the conditions under which the deployment of an additional robot may be required (e.g. based on field size and insect species). Additionally, the efficacy of the algorithm when multiple robots are operating simultaneously in the field should be scrutinized. Overall, these areas of inquiry could help to further refine and improve the proposed algorithm, ultimately leading to more effective insect detection and monitoring in agricultural settings.

# Appendices

## Appendix A- Algorithm without prior information

### A.1. Scenario A- Results

Scenario A- Simulation Input:

LOR = 100 (Length of rows)
WOR = 3.5 (Width of rows)
days = 3 (number of working days)
DistPlants = 2 (Distance Between Plants)
Dunam = 2
robot speed km\h = 2.5
DiseaseSeverity = 0.3 (Chances of disease spreading)
VP time (1: 4.34, 2: 16.14, 3: 23.74)
seconds turning of the robot (90®: 10/180®:23)
DiseaseProbability = 0.3
DiseaseSeverity = 0.3
totalTime = 7,200 (Total seconds robot works during the day- 2 hours)
timeBetweeenRows = 25 seconds

| Algorithm | Strategy | Num of plants between samples | PVV_ED1 | PCD_ED1 | PVV_ED2 | PCD_ED2 | PVV_ED3 | PCD_ED3 | Total visited plants % | Num of insected / detected / not detected | % of clear plants |
|---|---|---|---|---|---|---|---|---|---|---|---|
| 1 | samples every vine - Snake Strategy | 1 | 53 | 51.1 | 47 | 96.4 | 52.9 | 100 | 100 | 239/239/0 | 100 |
| 2 | samples every n plants - Snake Strategy | 2 | 53.8 | 50.5 | 46.2 | 60.5 | 53.6 | 44.6 | 100 | 332/223/109 | 78.2 |
| 2 | samples every n plants - Snake Strategy | 3 | 52.8 | 53.7 | 52.6 | 65.6 | 52.4 | 50.9 | 100 | 334/240/94 | 81.2 |
| 2 | samples every n plants - Snake Strategy | 4 | 52.2 | 52.6 | 52.2 | 69 | 52.2 | 51.8 | 100 | 353/268/85 | 83 |
| 3 | samples random plants- Snake Strategy | 1-4 | 53.8 | 52.1 | 45.2 | 41 | 46.8 | 53 | 82.2 | 320/222/98 | 80.4 |

| | | | | | | | | | | |
|---|---|---|---|---|---|---|---|---|---|---|
| 6 | samples every vine - Neighbor strategy | 1 | 82 | 79.1 | 82 | 100 | 80.1 | - | 100 | 151/151/0 | 100 |
| 4 | samples every n plants - Neighbor strategy | 2 | 80.8 | 83.3 | 80.8 | 100 | 80.8 | - | 100 | 154/154/0 | 100 |
| 4 | samples every n plants - Neighbor strategy | 3 | 80 | 84.1 | 66.8 | 80.1 | 73.6 | 85.7 | 100 | 186/184/2 | 99.6 |
| 4 | samples every n plants - Neighbor strategy | 4 | 78.8 | 82.1 | 78.4 | 83.6 | 78.4 | 81.8 | 100 | 225/219/6 | 98.8 |
| 5 | samples random plants- Neighbor strategy | 1-4 | 68 | 67.7 | 63.6 | 59.6 | 62.8 | 71.8 | 94.8 | 275/242/33 | 93.4 |

*Table 2: Scenario A- Performance measure results*

## A.2. Scenario B- Results

Scenario B- Vineyard 2 times larger:

LOR = **200** (Length of rows)
WOR = 3.5 (Width of rows)
days = 3 (number of working days)
DistPlants = **3** (Distance Between Plants)
Dunam = **4**
robot speed km\h = 2.5
DiseaseSeverity = 0.3 (Chances of disease spreading)
VP time (1: 4.34, 2: 16.14, 3: 23.74)
seconds turning of the robot (90®: 10/180®:23)
DiseaseProbability = 0.3
DiseaseSeverity = 0.3
totalTime = 7,200 (Total seconds robot works during the day- 2 hours)
timeBetweeenRows = 25 seconds

| Algorithm | Strategy | Num of plants between samples | PVV_ED1 | PCD_ED1 | PVV_ED2 | PCD_ED2 | PVV_ED3 | PCD_ED3 | Total visited plants % | Num of insected / detected / not detected | % of clear plants |
|---|---|---|---|---|---|---|---|---|---|---|---|
| 1 | samples every vine - Snake Strategy | 1 | 53 | 55.5 | 47 | 97.8 | 52.8 | 100 | 100 | 253/253/0 | 100 |
| 2 | samples every n plants - Snake Strategy | 2 | 37.3 | 35.7 | 37.3 | 44.4 | 25.3 | 40.9 | 100 | 259/266/193 | 71.2 |
| 2 | samples every n plants - Snake Strategy | 3 | 36.8 | 35.3 | 36.7 | 41.3 | 36.8 | 44 | 100 | 521/292/229 | 65.8 |
| 2 | samples every n plants - Snake Strategy | 4 | 37.6 | 41.2 | 37.5 | 40.4 | 36.8 | 46 | 100 | 218/292/226 | 66.2 |
| 3 | samples random plants - Snake Strategy | 1-4 | 37.3 | 34.7 | 34.9 | 37.9 | 33.3 | 34.3 | 65.2 | 510/203/307 | 54.2 |
| 6 | samples every vine - Neighbor strategy | 1 | 59.1 | 56.7 | 40.8 | 98.8 | 58.8 | 100 | 100 | 394/394/0 | 100 |
| 4 | samples every n plants - Neighbor strategy | 2 | 58.5 | 57.6 | 58.2 | 77.5 | 58.2 | 89.5 | 100 | 387/377/10 | 98.5 |
| 4 | samples every n plants - Neighbor strategy | 3 | 57.9 | 58.1 | 57.6 | 73.9 | 57.9 | 73.4 | 100 | 353/315/38 | 94.3 |
| 4 | samples every n plants - Neighbor strategy | 4 | 57.3 | 57.6 | 57.3 | 70.5 | 57.1 | 70.8 | 100 | 392/348/44 | 93.4 |
| 5 | samples random plants - Neighbor strategy | 1-4 | 49.8 | 49.5 | 48.6 | 35.5 | 48.3 | 51.8 | 81.5 | 481/318/163 | 75.6 |

*Table 3: Scenario B- Performance measure results*

## A.3. Scenario C- Results

Scenario C- Vineyard 4 times larger:

LOR = **400** (Length of rows)  
WOR = 3.5 (Width of rows)  
days = 3 (number of working days)  
DistPlants = **4** (Distance Between Plants)  
Dunam = **8**  
robot speed km\h = 2.5  
DiseaseSeverity = 0.3 (Chances of disease spreading)  
VP time (1: 4.34, 2: 16.14, 3: 23.74)  
seconds turning of the robot (90®: 10/180®:23)  
DiseaseProbability = 0.3  
DiseaseSeverity = 0.3  
totalTime = 7,200 (Total seconds robot works during the day- 2 hours)  
timeBetweeenRows = 25 seconds  

| Algorithm | Strategy | Num of plants between samples | PVV_ED1 | PCD_ED1 | PVV_ED2 | PCD_ED2 | PVV_ED3 | PCD_ED3 | Total visited plants % | Num of insected / detected / not detected | % of clear plants |
|---|---|---|---|---|---|---|---|---|---|---|---|
| 1 | samples every vine - Snake Strategy | 1 | 24.2 | 23.8 | 24.2 | 29.7 | 24.2 | 43.2 | 72.6 | 662/392/270 | 73 |
| 2 | samples every n plants - Snake Strategy | 2 | 24.4 | 24.1 | 24.4 | 31.1 | 24.1 | 32.9 | 72.7 | 683/271/412 | 58.9 |
| 2 | samples every n plants - Snake Strategy | 3 | 23.8 | 24.6 | 23.8 | 26.4 | 23.8 | 29.1 | 71.4 | 731/222/509 | 50.1 |
| 2 | samples every n plants - Snake Strategy | 4 | 24.4 | 24.7 | 23.5 | 25.5 | 24.3 | 31.1 | 72.2 | 728/231/497 | 50.3 |
| 3 | samples random plants - Snake Strategy | 1-4 | 24.2 | 23.2 | 23.9 | 24.5 | 22.1 | 19.5 | 61.6 | 669/171/498 | 48.4 |
| 6 | samples every vine - Neighbor strategy | 1 | 38.2 | 36.6 | 38 | 60.1 | 23.8 | 100 | 100 | 384/384/0 | 100 |
| 4 | samples every n plants - Neighbor strategy | 2 | 38 | 39.1 | 37.8 | 48.9 | 37.8 | 54.5 | 100 | 422/328/94 | 90.6 |
| 4 | samples every n plants - Neighbor strategy | 3 | 37.6 | 38.9 | 37.6 | 47.9 | 37.6 | 52.2 | 100 | 481/319/162 | 83.8 |
| 4 | samples every n plants - Neighbor strategy | 4 | 37.4 | 35.9 | 37.4 | 46.8 | 37.2 | 53.5 | 100 | 500/326/174 | 82.6 |
| 5 | samples random plants - Neighbor strategy | 1-4 | 38.2 | 43.7 | 31.4 | 46.4 | 30 | 4.6 | 71.4 | 438/215/223 | 77.4 |

*Table 4: Scenario C- Performance measure results*

## A.4. All Scenarios- Days to specific % detection

The table below provides a comparison of the performance of different algorithms in different scenarios, in terms of the number of days it takes to reach specific detection percentages (30%, 50%, 80%, 100%).

| Algorithm | Strategy | Num of plants between samples | Scenario A | | | | Scenario B | | | | Scenario C | | | |
|---|---|---|---|---|---|---|---|---|---|---|---|---|---|---|
| | | | D30 | D50 | D80 | D100 | D30 | D50 | D80 | D100 | D30 | D50 | D80 | D100 |
| 1 | samples every vine - Snake Strategy | 1 | 1 | 1 | 2 | 2 | 1 | 1 | 2 | 3 | 2 | 3 | - | - |
| 2 | samples every n plants - Snake Strategy | 2 | 1 | 1 | 2 | - | 1 | 2 | - | - | 2 | 3 | - | - |
| 2 | samples every n plants - Snake Strategy | 3 | 1 | 1 | 2 | - | 1 | 2 | - | - | 2 | 3 | - | - |
| 2 | samples every n plants - Snake Strategy | 4 | 1 | 1 | 3 | - | 1 | 2 | - | - | 2 | 3 | - | - |
| 3 | samples random plants- Snake Strategy | 1-4 | 1 | 2 | 3 | - | 1 | 2 | - | - | 2 | - | - | - |
| 6 | samples every vine - Neighbor strategy | 1 | 1 | 1 | 1 | 2 | 1 | 1 | 2 | 2 | 1 | 2 | 3 | 3 |

| 4 | samples every n plants - Neighbor strategy | 2 | 1 | 1 | 1 | 2 | 1 | 1 | 2 | - | 1 | 2 | 3 | - |
| 4 | samples every n plants - Neighbor strategy | 3 | 1 | 1 | 1 | 2 | 1 | 1 | 2 | - | 1 | 2 | 3 | - |
| 4 | samples every n plants - Neighbor strategy | 4 | 1 | 1 | 1 | 2 | 1 | 1 | 2 | - | 1 | 2 | 3 | - |
| 5 | samples random plants- Neighbor strategy | 1-4 | 1 | 1 | 2 | - | 1 | 1 | 2 | - | 1 | 2 | - | - |

*Table 5: Scenarios Comparison - Days to specific % detection*

### A.5. Matlab code- snake strategy

```
% initialize variables
LOR = 100; %Length of rows
WOR = 3.5; %Width of rows
days = 3; %number of days for the simulation
DistPlants = 2; %Distance Between Plants- meters
Dunam = 2 ;
Robot_Speed_KMH = 2.5; % Robot Speed - Kilometers per hour - average
Robot_Speed = (Robot_Speed_KMH*1000)/60/60; %Robot speed - Meters per Seconds
DiseaseProbability = 0.3; %chance of having a bug on the vine
DiseaseSeverity = 0.3; %defines how fast distributes to neighbor plants
NumOfPlantsInRow = int16(LOR/DistPlants); %calc number of plants in a single row
NumOfRows = int16(Dunam/(LOR*WOR)); %calc number of rows in the green house
NumOfRowsToCheck = int16(NumOfRows*2-2); %each row has two lines of plants- except of the first and the last.
oneVPTime = 4.34; %total seconds for one View Point
twoVPsTime = 16.14; %total seconds for two View Points
threeVPsTime = 23.74; %total seconds for three View Points
turningTime90 = 10; %total seconds for turning to the plants/back to road 90 degrees **Check the correct time**
turningTime180 = 23; %total seconds for turning 180 degrees  to the next
```

```matlab
totalTime = 2*60*60; %total second robot works during the day
timeLeft = totalTime; %total seconds left for the rest of the day
timeBetweenPlants = DistPlants/Robot_Speed; %navigating time between two adjacent plants- seconds
timeBetweeenRows = 25; %navigating time between two adjacent rows- seconds
numOfDetections = 0; %count number of Correct Detections all days
numOfDetectionsToday = 0;%count number of Correct Detections this day
TotalBugsToday = 0;%sum number of bugs at the begining of a day
DetectionPercent = [];
TimeDetectionPercent = []; %working time in Hours

Bugs_Matrix = zeros(NumOfPlantsInRow,NumOfRowsToCheck); %matrix for the bugs locations
Visited_Matrix = zeros(NumOfPlantsInRow,NumOfRowsToCheck);%presents which plants the robot visited
Visited_Today_Matrix = zeros(NumOfPlantsInRow,NumOfRowsToCheck);%presents which plants the robot visited Today

%adds Bugs to the matrix randomally
pos = 1; %index for the matrix cells
for i = 1:(NumOfRowsToCheck/2+1)
    for j = 1:NumOfPlantsInRow
        if (i == (NumOfRowsToCheck/2 + 1))%last row
            if rand > (1-DiseaseProbability) %chances for having a bug on the vine
                Bugs_Matrix(pos - NumOfPlantsInRow) = 1;
            else
                Bugs_Matrix(pos - NumOfPlantsInRow) = 0;
            end
        elseif (i == 1) %first row
            if rand > (1-DiseaseProbability) %chances for having a bug on the vine
                Bugs_Matrix(pos) = 1;
            else
                Bugs_Matrix(pos) = 0;
            end
        else %all the rows except the first and the last
            if rand > (1-DiseaseProbability) %chances for having a bug on the vine
                Bugs_Matrix(pos) = 1;
                Bugs_Matrix(pos - NumOfPlantsInRow) = 1; %adjacent plant
            else
                Bugs_Matrix(pos) = 0;
            end
        end
        pos = pos +1;
```

```matlab
        end
        pos = pos + NumOfPlantsInRow;
    end
    SumOfBugs = sum(Bugs_Matrix(:) == 1); %sum number of bugs

    %strategy 2 - sample every n plants + m VPs - moves as snake
    needToMoveBetweenRows = true; % indicates which side of the robot the plants are
    pos = 1; %index for the matrix cells
    forward = true; %indicates the direction the robot should go
    for h =1:days
        Visited_Today_Matrix = zeros(NumOfPlantsInRow,NumOfRowsToCheck);
        numOfDetectionsToday = 0;
        newBugsToday = 0;
        TotalBugsToday = sum(Bugs_Matrix(:) == 1);
        timeLeft = totalTime; %total seconds left for the rest of the day
        n = 4;
        VPtime = threeVPsTime;
        if(timeLeft-(timeBetweenPlants + VPtime + turningTime90) > 0)
            while(timeLeft-(timeBetweenPlants + VPtime + turningTime90) > 0); % there is more working time
                if(forward == true)
                    for i = 1:NumOfRowsToCheck;
                        if(pos + n > (NumOfRowsToCheck*NumOfPlantsInRow))%out of bounds
                            forward = false;
                            break
                        end
                        for j = 1:n:NumOfPlantsInRow;
                            if((timeLeft-(timeBetweenPlants + VPtime)) > 0); %working time hasn't finished yet, and the robot can check another vine
                                timeLeft = timeLeft - (timeBetweenPlants + VPtime);
                                Visited_Matrix(pos) = Visited_Matrix(pos)+ 1;
                                Visited_Today_Matrix(pos) = Visited_Today_Matrix(pos)+ 1;
                            if (Bugs_Matrix(pos) == 1); %in case of detection
                                Bugs_Matrix(pos) = 0;
                                numOfDetections = numOfDetections + 1;
                                numOfDetectionsToday = numOfDetectionsToday + 1;
                            end
                            if(j == NumOfPlantsInRow);%if it is the last vine in the row
                                if(needToMoveBetweenRows == true)%
                                    timeLeft = timeLeft - timeBetweeenRows; %calculate time between rows
                                else
```

```matlab
                    timeLeft = timeLeft - turningTime180; %calculate turning time at the same row
                end
            end
            else
                break
            end
            if(pos + n > (NumOfRowsToCheck*NumOfPlantsInRow))%out of bounds
                minvalue = min(Visited_Matrix(NumOfPlantsInRow-n+1:NumOfPlantsInRow,NumOfRowsToCheck));
                for(k = NumOfPlantsInRow - n + 1:NumOfPlantsInRow)
                    if(Visited_Matrix((NumOfRowsToCheck - 1) * NumOfPlantsInRow + k) == minvalue)
                        pos = (NumOfRowsToCheck - 1) * NumOfPlantsInRow + k;
                    end
                end
                forward = false;
                break
            else
                pos = pos + n;
            end
            DetectionPercent (end+1) = numOfDetections/SumOfBugs * 100;
            TimeDetectionPercent (end+1) = (totalTime*h - timeLeft)/60/60; %working time in Hours
        end
    end
    else
        for i = NumOfRowsToCheck:-1:1;
            if(pos - n < 1) %out of bounds
                forward = true;
                break
            end
            for j = NumOfPlantsInRow:-n:1;
                if((timeLeft-(timeBetweenPlants + VPtime)) > 0); %working time hasn't finished yet, and the robot can check another vine
                    timeLeft = timeLeft - (timeBetweenPlants + VPtime);
                    Visited_Matrix(pos) = Visited_Matrix(pos)+ 1;
                    Visited_Today_Matrix(pos) = Visited_Today_Matrix(pos)+ 1;
                    if (Bugs_Matrix(pos) == 1); %in case of detection
                        Bugs_Matrix(pos) = 0;
                        numOfDetections = numOfDetections + 1;
                        numOfDetectionsToday = numOfDetectionsToday + 1;
                    end
```

```matlab
                    if(j == NumOfPlantsInRow);%if it is the last vine in the row
                        if(needToMoveBetweenRows == true)
                            timeLeft = timeLeft - timeBetweeenRows; %calculate time between rows
                        else
                            timeLeft = timeLeft - turningTime180; %calculate turning time at the same row
                        end
                    end
                else
                    break
                end
                if(pos - n < 1) %out of bounds
                    minvalue = min(Visited_Matrix(1:n,1));
                    for(k = 1 : n )
                        if(Visited_Matrix(k) == minvalue)
                            pos = k;
                        end
                    end
                    forward = false;
                    break
                else
                    pos = pos - n;
                end
                DetectionPercent (end+1) = numOfDetections/SumOfBugs * 100;
                TimeDetectionPercent (end+1) = (totalTime*h - timeLeft)/60/60; %working time in Hours
            end

        end
       end
      end
   end
   VisitedToday = sum(Visited_Today_Matrix(:) ~= 0) / double(NumOfRowsToCheck*NumOfPlantsInRow) *100;
   VisitedTodayToPrint = ['Visited Percentage of day number ',num2str(h), ' is: ' , num2str(VisitedToday)];
   disp (VisitedTodayToPrint)
   DetectionPercentageToday = numOfDetectionsToday/TotalBugsToday*100;
   DetectionPercentageToPrint = ['Detection Percentage of day number ',num2str(h), ' is: ' , num2str(DetectionPercentageToday)];
   disp(DetectionPercentageToPrint)
   pos_Bugs_Matrix = 1;
   Helper_Bugs_Matrix = Bugs_Matrix;
```

```
rowIndex = 1 ; %index for which row
left = true; % it has neighbor at the right side
while (pos_Bugs_Matrix < NumOfPlantsInRow * NumOfRowsToCheck + 1)
   if (rowIndex == 1 ) % checking first row
      if(pos_Bugs_Matrix == 1) % first vine first row
         if Helper_Bugs_Matrix(pos_Bugs_Matrix) == 1 && rand < DiseaseSeverity && Bugs_Matrix(pos_Bugs_Matrix + 1) ~= 1  %chance of spreading the disease
            Bugs_Matrix(pos_Bugs_Matrix + 1) = 1;
            newBugsToday = newBugsToday + 1;
         end
      elseif (pos_Bugs_Matrix == NumOfPlantsInRow ) %last vine first row
         rowIndex = 2;
         if (rand < DiseaseSeverity && Helper_Bugs_Matrix(pos_Bugs_Matrix) == 1 && Bugs_Matrix(pos_Bugs_Matrix - 1) ~= 1) %chance of spreading the disease
            Bugs_Matrix(pos_Bugs_Matrix - 1) = 1;
            newBugsToday = newBugsToday + 1;
         end
      else %first row
         if (Helper_Bugs_Matrix(pos_Bugs_Matrix) == 1)
            if rand < DiseaseSeverity %chance of spreading the disease
               if Bugs_Matrix(pos_Bugs_Matrix - 1) ~= 1
                  Bugs_Matrix(pos_Bugs_Matrix - 1) = 1;
                  newBugsToday = newBugsToday + 1;
               end
               if Bugs_Matrix(pos_Bugs_Matrix + 1) ~= 1
                  Bugs_Matrix(pos_Bugs_Matrix + 1) = 1;
                  newBugsToday = newBugsToday + 1;
               end
            end
         end
      end
   elseif (rowIndex == NumOfRowsToCheck ) % checking last row
      if((pos_Bugs_Matrix == NumOfPlantsInRow * (NumOfRowsToCheck - 1 ) + 1)) % first vine last row
         if rand < DiseaseSeverity && Bugs_Matrix(pos_Bugs_Matrix + 1) ~= 1 && Helper_Bugs_Matrix(pos_Bugs_Matrix) == 1 %chance of spreading the disease
            Bugs_Matrix(pos_Bugs_Matrix + 1) = 1;
            newBugsToday = newBugsToday + 1;
         end
      elseif (pos_Bugs_Matrix == NumOfPlantsInRow * NumOfRowsToCheck ) %last vine last row
```

```matlab
                if rand < DiseaseSeverity && Bugs_Matrix(pos_Bugs_Matrix - 1) ~= 1 && Helper_Bugs_Matrix(pos_Bugs_Matrix) == 1%chance of spreading the disease
                    Bugs_Matrix(pos_Bugs_Matrix - 1) = 1;
                    newBugsToday = newBugsToday + 1;
                end
            else %last row
                if (Helper_Bugs_Matrix(pos_Bugs_Matrix) == 1)
                    if rand < DiseaseSeverity %chance of spreading the disease
                        if Bugs_Matrix(pos_Bugs_Matrix - 1) ~= 1
                            Bugs_Matrix(pos_Bugs_Matrix - 1) = 1;
                            newBugsToday = newBugsToday + 1;
                        end
                        if Bugs_Matrix(pos_Bugs_Matrix + 1) ~= 1
                            Bugs_Matrix(pos_Bugs_Matrix + 1) = 1;
                            newBugsToday = newBugsToday + 1;
                        end
                    end
                end
            end
        else % rows with neighbor
            if(pos_Bugs_Matrix - (NumOfPlantsInRow * (rowIndex - 1)) == 1 ) % first vine in the row
                if rand < DiseaseSeverity && Helper_Bugs_Matrix(pos_Bugs_Matrix) == 1 %chance of spreading the disease
                    if left == true
                        if Bugs_Matrix(pos_Bugs_Matrix + 1) ~= 1
                            Bugs_Matrix(pos_Bugs_Matrix + 1) = 1;  % neigbhor from the right
                            newBugsToday = newBugsToday + 1;
                        end
                        if Bugs_Matrix(pos_Bugs_Matrix + NumOfPlantsInRow) ~= 1
                            Bugs_Matrix(pos_Bugs_Matrix + NumOfPlantsInRow) = 1;
                            newBugsToday = newBugsToday + 1;
                        end
                    else
                        if Bugs_Matrix(pos_Bugs_Matrix + 1) ~= 1
                            Bugs_Matrix(pos_Bugs_Matrix + 1) = 1;  % neigbhor from the left
                            newBugsToday = newBugsToday + 1;
                        end
                        if Bugs_Matrix(pos_Bugs_Matrix - NumOfPlantsInRow) ~= 1
                            Bugs_Matrix(pos_Bugs_Matrix - NumOfPlantsInRow) = 1;
                            newBugsToday = newBugsToday + 1;
```

```matlab
                    end
                end
            end
        elseif(pos_Bugs_Matrix - (NumOfPlantsInRow * (rowIndex - 1)) == NumOfPlantsInRow) %last vine in the row
            rowIndex = rowIndex + 1;
              if left == true
                left = false;
                    if(Helper_Bugs_Matrix(pos_Bugs_Matrix) == 1 && rand < DiseaseSeverity)
                        if Bugs_Matrix(pos_Bugs_Matrix - 1) ~= 1
                            Bugs_Matrix(pos_Bugs_Matrix - 1) = 1;  % neigbhor from the right
                            newBugsToday = newBugsToday + 1;
                        end
                        if Bugs_Matrix(pos_Bugs_Matrix + NumOfPlantsInRow) ~= 1
                            Bugs_Matrix(pos_Bugs_Matrix + NumOfPlantsInRow) = 1;
                            newBugsToday = newBugsToday + 1;
                        end
                    end
              else
                left = true;
                if(Helper_Bugs_Matrix(pos_Bugs_Matrix) == 1 && rand < DiseaseSeverity)
                    if Bugs_Matrix(pos_Bugs_Matrix - 1) ~= 1
                        Bugs_Matrix(pos_Bugs_Matrix - 1) = 1;  % neigbhor from the left
                        newBugsToday = newBugsToday + 1;
                    end
                    if Bugs_Matrix(pos_Bugs_Matrix - NumOfPlantsInRow) ~= 1
                        Bugs_Matrix(pos_Bugs_Matrix - NumOfPlantsInRow) = 1;
                        newBugsToday = newBugsToday + 1;
                    end
                 end
                end
        else
            if (rand < DiseaseSeverity && Helper_Bugs_Matrix(pos_Bugs_Matrix) == 1) %chance of spreading the disease
                if left == true
                    if Bugs_Matrix(pos_Bugs_Matrix - 1) ~= 1
                        Bugs_Matrix(pos_Bugs_Matrix - 1) = 1;
                        newBugsToday = newBugsToday + 1;
                    end
                    if Bugs_Matrix(pos_Bugs_Matrix + 1) ~= 1
```

```
                    Bugs_Matrix(pos_Bugs_Matrix + 1) = 1;
                    newBugsToday = newBugsToday + 1;
                  end
                  if Bugs_Matrix(pos_Bugs_Matrix + NumOfPlantsInRow) ~= 1
                    Bugs_Matrix(pos_Bugs_Matrix + 1) = 1;
                    newBugsToday = newBugsToday + 1;
                  end
                else
                  if Bugs_Matrix(pos_Bugs_Matrix - 1) ~= 1
                    Bugs_Matrix(pos_Bugs_Matrix - 1) = 1;
                    newBugsToday = newBugsToday + 1;
                  end
                  if Bugs_Matrix(pos_Bugs_Matrix + 1) ~= 1
                    Bugs_Matrix(pos_Bugs_Matrix + 1) = 1;
                    newBugsToday = newBugsToday + 1;
                  end
                  if Bugs_Matrix(pos_Bugs_Matrix - NumOfPlantsInRow) ~= 1
                    Bugs_Matrix(pos_Bugs_Matrix - NumOfPlantsInRow) = 1;
                    newBugsToday = newBugsToday + 1;
                  end
                end
              end
            end
          end
          pos_Bugs_Matrix = pos_Bugs_Matrix +1;
        end
        if h~=days
            SumOfBugs = SumOfBugs + newBugsToday;
        end
end

DetectionPercent (end+1) = numOfDetections/SumOfBugs * 100;
TimeDetectionPercent (end+1) = (totalTime*h - timeLeft)/60/60; %working time in Hours

%performance
clearSpots = sum(Bugs_Matrix(:) == 0);
PercentOfDetections = clearSpots/double(NumOfRowsToCheck*NumOfPlantsInRow)
VisitedPercentageAllDays = sum(Visited_Matrix(:) ~= 0) / double(NumOfRowsToCheck*NumOfPlantsInRow)
*100 %percentage of visited plants
NumOfBugshaventFound =  sum(Bugs_Matrix(:) == 1) %Num Of Bugas that havent found yet
```

plot(TimeDetectionPercent/2,DetectionPercent)

title('Detection Percent: Snake Strategy');

xlabel('Days');

ylabel('Percent detection of infected plants');

### A.6. Matlab Code- Neighbor Strategy

```
% initialize variables
LOR = 100; %Length of rows
WOR = 3.5; %Width of rows
days = 3; %number of days for the simulation
DistPlants = 2; %Distance Between Plants- meters
Dunam = 2 ;
Robot_Speed_KMH = 2.5; % Robot Speed - Kilometers per hour - average
Robot_Speed = (Robot_Speed_KMH*1000)/60/60; %Robot speed - Meters per Seconds
DiseaseProbability = 0.3; %chance of having a bug on the vine
DiseaseSeverity = 0.9; %defines how fast distributes to neighbor plants
NumOfPlantsInRow = int16(LOR/DistPlants); %calc number of plants in a single row
NumOfRows = int16(Dunam/(LOR*WOR)); %calc number of rows in the green house
NumOfRowsToCheck = int16(NumOfRows*2-2); %each row has two lines of plants- except of the first and the last.
oneVPTime = 4.34; %total seconds for one View Point
twoVPsTime = 16.14; %total seconds for two View Points
threeVPsTime = 23.74; %total seconds for three View Points
turningTime90 = 10; %total seconds for turning to the plants/back to road 90 degrees **Check the correct time**
turningTime180 = 23; %total seconds for turning 180 degrees  to the next
totalTime = 2*60*60; %total second robot works during the day
timeLeft = totalTime; %total seconds left for the rest of the day
timeBetweenPlants = DistPlants/Robot_Speed; %navigating time between two adjacent plants- seconds
timeBetweeenRows = 25; %navigating time between two adjacent rows- seconds
numOfDetections = 0; %count number of Correct Detections all days
numOfDetectionsToday = 0;%count number of Correct Detections this day
TotalBugsToday = 0;%sum number of bugs at the begining of a day
newBugsToday = 0; %count number of new bugs today
DetectionPercent = []; %List of detection percent - function of time
TimeDetectionPercent = []; %List of detection percent - function of time

Bugs_Matrix = zeros(NumOfPlantsInRow,NumOfRowsToCheck); %matrix for the bugs locations
Visited_Matrix = zeros(NumOfPlantsInRow,NumOfRowsToCheck);%presents which plants the robot visited
Visited_Today_Matrix = zeros(NumOfPlantsInRow,NumOfRowsToCheck);%presents which plants the robot visited Today
```

```matlab
%adds Bugs to the matrix randomally
pos = 1; %index for the matrix cells
for i = 1:(NumOfRowsToCheck/2+1)
    for j = 1:NumOfPlantsInRow
        if (i == (NumOfRowsToCheck/2 + 1))%last row
            if rand > (1-DiseaseProbability) %chances for having a bug on the vine
                Bugs_Matrix(pos - NumOfPlantsInRow) = 1;
            else
                Bugs_Matrix(pos - NumOfPlantsInRow) = 0;
            end
        elseif (i == 1) %first row
            if rand > (1-DiseaseProbability) %chances for having a bug on the vine
                Bugs_Matrix(pos) = 1;
            else
                Bugs_Matrix(pos) = 0;
            end
        else %all the rows except the first and the last
            if rand > (1-DiseaseProbability) %chances for having a bug on the vine
                Bugs_Matrix(pos) = 1;
                Bugs_Matrix(pos - NumOfPlantsInRow) = 1; %adjacent plant
            else
                Bugs_Matrix(pos) = 0;
            end
        end
        pos = pos +1;
    end
    pos = pos + NumOfPlantsInRow;
end
SumOfBugs = sum(Bugs_Matrix(:) == 1) %sum number of bugs

%strategy 1 - sample every n plants + m VPs - adjacent plants
pos = 1; %index for the matrix cells
forward = true; %indicates the direction the robot should go
RowIndex = 1; %indicates which row the robot is currently at
for h = 1:days
    newBugsToday = 0;
    Visited_Today_Matrix = zeros(NumOfPlantsInRow,NumOfRowsToCheck);
    numOfDetectionsToday = 0;
    TotalBugsToday = sum(Bugs_Matrix(:) == 1);
```

```matlab
    timeLeft = totalTime; %total seconds left for the rest of the day
    n = 4;
    VPtime = oneVPTime;
    while(timeLeft-(timeBetweenPlants + VPtime * 2 +turningTime180) > 0) % there is more working time
        if(forward == true)
            while (pos + NumOfPlantsInRow < NumOfPlantsInRow * NumOfRowsToCheck + 1)
                if((timeLeft-(timeBetweenPlants + VPtime * 2 +turningTime180)) > 0) %working time hasn't finished yet, and the robot can check another vine
                    timeLeft = timeLeft - (timeBetweenPlants + VPtime * 2 +turningTime180);
                    Visited_Matrix(pos) = Visited_Matrix(pos) + 1;
                    Visited_Matrix(pos + NumOfPlantsInRow) = Visited_Matrix(pos + NumOfPlantsInRow)+ 1;
                    Visited_Today_Matrix(pos) = Visited_Today_Matrix(pos) + 1;
                    Visited_Today_Matrix(pos + NumOfPlantsInRow) = Visited_Today_Matrix(pos + NumOfPlantsInRow)+ 1;
                    if (Bugs_Matrix(pos) == 1 ) %in case of detection
                        Bugs_Matrix(pos) = 0;
                        numOfDetections = numOfDetections + 1;
                        numOfDetectionsToday = numOfDetectionsToday + 1;
                    end
                    if (Bugs_Matrix(pos + NumOfPlantsInRow) == 1 ) %in case of detection
                        Bugs_Matrix(pos + NumOfPlantsInRow) = 0;
                        numOfDetections = numOfDetections + 1;
                        numOfDetectionsToday = numOfDetectionsToday + 1;
                    end
                else
                    break
                end
                if (pos + n <= RowIndex * NumOfPlantsInRow) % checks if we moved between rows
                    pos = pos + n;
                else %we moved between rows
                    pos = pos + n + NumOfPlantsInRow;
                    timeLeft = timeLeft - timeBetweeenRows; %calculate time between rows
                    RowIndex = RowIndex + 2;
                end
                if(pos > NumOfPlantsInRow * NumOfRowsToCheck)
                    forward = false;
                    minvalue = min(Visited_Matrix(NumOfPlantsInRow - n + 1:NumOfPlantsInRow,NumOfRowsToCheck));
                    for k = NumOfPlantsInRow - n + 1:NumOfPlantsInRow
                        index = NumOfPlantsInRow * NumOfRowsToCheck - NumOfPlantsInRow + k;
```

```matlab
                    if(Visited_Matrix(index) == minvalue)
                        pos = index;
                    end
                end
                break
            end
                DetectionPercent (end+1) = numOfDetections/SumOfBugs * 100;
                TimeDetectionPercent (end+1) = (totalTime*h - timeLeft)/60/60; %working time in Hours
        end
    else
        while (pos - NumOfPlantsInRow > 0)
            if((timeLeft-(timeBetweenPlants + VPtime * 2 +turningTime180)) > 0) %working time hasn't finished yet, and the robot can check another vine
                timeLeft = timeLeft - (timeBetweenPlants + VPtime * 2 +turningTime180);
                Visited_Matrix(pos) = Visited_Matrix(pos) + 1;
                Visited_Matrix(pos - NumOfPlantsInRow) = Visited_Matrix(pos - NumOfPlantsInRow)+ 1;
                Visited_Today_Matrix(pos) = Visited_Today_Matrix(pos) + 1;
                Visited_Today_Matrix(pos - NumOfPlantsInRow) = Visited_Today_Matrix(pos - NumOfPlantsInRow)+ 1;
                if (Bugs_Matrix(pos) == 1 ) %in case of detection
                    Bugs_Matrix(pos) = 0;
                    numOfDetections = numOfDetections + 1;
                    numOfDetectionsToday = numOfDetectionsToday + 1;
                end
                if (Bugs_Matrix(pos - NumOfPlantsInRow) == 1 ) %in case of detection
                    Bugs_Matrix(pos - NumOfPlantsInRow) = 0;
                    numOfDetections = numOfDetections + 1;
                    numOfDetectionsToday = numOfDetectionsToday + 1;
                end
            else
                break
            end
            if (pos - n > (RowIndex-2) * NumOfPlantsInRow)% checks if we moved between rows
                pos = pos - n;
            else %we moved between rows
                pos = pos - n - NumOfPlantsInRow;
                timeLeft = timeLeft - timeBetweeenRows; %calculate time between rows
                RowIndex = RowIndex - 2;
            end
            if(pos - NumOfPlantsInRow < 1)
```

```
                forward = true;
                minvalue = min(Visited_Matrix(1:n,1));
                for(k = n:-1:1)
                    if(Visited_Matrix(k) == minvalue)
                        pos = k;
                    end
                end
                break
            end
                DetectionPercent (end+1) = numOfDetections/SumOfBugs * 100;
                TimeDetectionPercent (end+1) = (totalTime*h - timeLeft)/60/60; %working time in Hours
        end
      end
    end
    VisitedToday = sum(Visited_Today_Matrix(:) ~= 0) / double(NumOfRowsToCheck*NumOfPlantsInRow) *100;
    VisitedTodayToPrint = ['Visited Percentage of day number ',num2str(h), ' is: ' , num2str(VisitedToday)];
    disp (VisitedTodayToPrint)
    DetectionPercentageToday = numOfDetectionsToday/TotalBugsToday*100;
    DetectionPercentageToPrint = ['Detection Percentage of day number ',num2str(h), ' is: ' , num2str(DetectionPercentageToday)];
    disp(DetectionPercentageToPrint)
    pos_Bugs_Matrix = 1;
    Helper_Bugs_Matrix = Bugs_Matrix;
    rowIndex = 1 ; %index for which row
    left = true; % it has neighbor at the right side
    while (pos_Bugs_Matrix < NumOfPlantsInRow * NumOfRowsToCheck + 1)
        if (rowIndex == 1 ) % checking first row
            if(pos_Bugs_Matrix == 1 && Helper_Bugs_Matrix(pos_Bugs_Matrix) == 1) % first vine first row
                if rand < DiseaseSeverity %chance of spreading the disease
                    Bugs_Matrix(pos_Bugs_Matrix + 1) = 1;
                    newBugsToday = newBugsToday + 1;
                end
            elseif (pos_Bugs_Matrix == NumOfPlantsInRow && Helper_Bugs_Matrix(pos_Bugs_Matrix) == 1) %last vine first row
                if rand < DiseaseSeverity %chance of spreading the disease
                    Bugs_Matrix(pos_Bugs_Matrix - 1) = 1;
                    newBugsToday = newBugsToday + 1;
                    rowIndex = 2;
                end
```

```matlab
            else %first row
                if (Helper_Bugs_Matrix(pos_Bugs_Matrix) == 1)
                    if rand < DiseaseSeverity %chance of spreading the disease
                        if  Bugs_Matrix(pos_Bugs_Matrix - 1) ~= 1
                            Bugs_Matrix(pos_Bugs_Matrix - 1) = 1;
                            newBugsToday = newBugsToday + 1;
                        end
                        if Bugs_Matrix(pos_Bugs_Matrix + 1) ~= 1
                            Bugs_Matrix(pos_Bugs_Matrix + 1) = 1;
                            newBugsToday = newBugsToday + 1;
                        end
                    end
                end
            end
        elseif (rowIndex == NumOfRowsToCheck ) % checking last row
            if(pos_Bugs_Matrix    ==    NumOfPlantsInRow  *  (NumOfRowsToCheck  -  1  )  +  1  && Helper_Bugs_Matrix(pos_Bugs_Matrix) == 1) % first vine last row
                if rand < DiseaseSeverity && Bugs_Matrix(pos_Bugs_Matrix + 1) ~= 1 %chance of spreading the disease
                    Bugs_Matrix(pos_Bugs_Matrix + 1) = 1;
                    newBugsToday = newBugsToday + 1;
                end
            elseif    (pos_Bugs_Matrix     ==     NumOfPlantsInRow     *     NumOfRowsToCheck     && Helper_Bugs_Matrix(pos_Bugs_Matrix) == 1) %last vine first row
                if rand < DiseaseSeverity && Bugs_Matrix(pos_Bugs_Matrix - 1) ~= 1 %chance of spreading the disease
                    Bugs_Matrix(pos_Bugs_Matrix - 1) = 1;
                    newBugsToday = newBugsToday + 1;
                end
            else %last row
                if (Helper_Bugs_Matrix(pos_Bugs_Matrix) == 1)
                    if rand < DiseaseSeverity %chance of spreading the disease
                        if Bugs_Matrix(pos_Bugs_Matrix - 1) ~= 1
                            Bugs_Matrix(pos_Bugs_Matrix - 1) = 1;
                            newBugsToday = newBugsToday + 1;
                        end
                        if Bugs_Matrix(pos_Bugs_Matrix + 1) ~= 1
                            Bugs_Matrix(pos_Bugs_Matrix + 1) = 1;
                            newBugsToday = newBugsToday + 1;
                        end
```

```matlab
                end
            end
        end
    else % rows with neighbor
        if(pos_Bugs_Matrix - (NumOfPlantsInRow * (rowIndex - 1)) == 1 && Helper_Bugs_Matrix(pos_Bugs_Matrix) == 1) % first vine in the row
            if rand < DiseaseSeverity %chance of spreading the disease
                if left == true
                    if Bugs_Matrix(pos_Bugs_Matrix + 1) ~= 1
                        Bugs_Matrix(pos_Bugs_Matrix + 1) = 1;  % neigbhor from the right
                        newBugsToday = newBugsToday + 1;
                    end
                    if Bugs_Matrix(pos_Bugs_Matrix + NumOfPlantsInRow) ~= 1
                        Bugs_Matrix(pos_Bugs_Matrix + NumOfPlantsInRow) = 1;
                        newBugsToday = newBugsToday + 1;
                    end
                else
                    if Bugs_Matrix(pos_Bugs_Matrix + 1) ~= 1
                        Bugs_Matrix(pos_Bugs_Matrix + 1) = 1;  % neigbhor from the left
                        newBugsToday = newBugsToday + 1;
                    end
                    if Bugs_Matrix(pos_Bugs_Matrix - NumOfPlantsInRow) ~= 1
                        Bugs_Matrix(pos_Bugs_Matrix - NumOfPlantsInRow) = 1;
                        newBugsToday = newBugsToday + 1;
                    end
                end
            end
        elseif(pos_Bugs_Matrix - (NumOfPlantsInRow * (rowIndex - 1)) == NumOfPlantsInRow) %last vine in the row
            if left == true
                left = false;
                if(Helper_Bugs_Matrix(pos_Bugs_Matrix) == 1 && rand < DiseaseSeverity)
                    if Bugs_Matrix(pos_Bugs_Matrix - 1) ~= 1
                        Bugs_Matrix(pos_Bugs_Matrix - 1) = 1;  % neigbhor from the right
                        newBugsToday = newBugsToday + 1;
                    end
                    if Bugs_Matrix(pos_Bugs_Matrix + NumOfPlantsInRow) ~= 1
                        Bugs_Matrix(pos_Bugs_Matrix + NumOfPlantsInRow) = 1;
                        newBugsToday = newBugsToday + 1;
                    end
```

```matlab
                    end
                else
                    left = true;
                    if(Helper_Bugs_Matrix(pos_Bugs_Matrix) == 1 && rand < DiseaseSeverity)
                        if Bugs_Matrix(pos_Bugs_Matrix - 1) ~= 1
                            Bugs_Matrix(pos_Bugs_Matrix - 1) = 1;  % neigbhor from the left
                            newBugsToday = newBugsToday + 1;
                        end
                        if Bugs_Matrix(pos_Bugs_Matrix - NumOfPlantsInRow) ~= 1
                            Bugs_Matrix(pos_Bugs_Matrix - NumOfPlantsInRow) = 1;
                            newBugsToday = newBugsToday + 1;
                        end
                    end
                end
            else
                if (rand < DiseaseSeverity && Helper_Bugs_Matrix(pos_Bugs_Matrix) == 1) %chance of spreading the disease
                    if left == true
                        if Bugs_Matrix(pos_Bugs_Matrix - 1) ~= 1
                            Bugs_Matrix(pos_Bugs_Matrix - 1) = 1;
                            newBugsToday = newBugsToday + 1;
                        end
                        if Bugs_Matrix(pos_Bugs_Matrix + 1) ~= 1
                            Bugs_Matrix(pos_Bugs_Matrix + 1) = 1;
                            newBugsToday = newBugsToday + 1;
                        end
                        if Bugs_Matrix(pos_Bugs_Matrix + NumOfPlantsInRow) ~= 1
                            Bugs_Matrix(pos_Bugs_Matrix + 1) = 1;
                            newBugsToday = newBugsToday + 1;
                        end
                    else
                        if Bugs_Matrix(pos_Bugs_Matrix - 1) ~= 1
                            Bugs_Matrix(pos_Bugs_Matrix - 1) = 1;
                            newBugsToday = newBugsToday + 1;
                        end
                        if Bugs_Matrix(pos_Bugs_Matrix + 1) ~= 1
                            Bugs_Matrix(pos_Bugs_Matrix + 1) = 1;
                            newBugsToday = newBugsToday + 1;
                        end
                        if Bugs_Matrix(pos_Bugs_Matrix - NumOfPlantsInRow) ~= 1
```

```
                        Bugs_Matrix(pos_Bugs_Matrix - NumOfPlantsInRow) = 1;
                        newBugsToday = newBugsToday + 1;
                    end
                end
            end
        end
    end
    pos_Bugs_Matrix = pos_Bugs_Matrix +1;
  end
  if h~=days
     SumOfBugs = SumOfBugs + newBugsToday;
  end
end

%performance
clearSpots = sum(Bugs_Matrix(:) == 0);
PercentOfDetections = clearSpots/double(NumOfRowsToCheck*NumOfPlantsInRow)
VisitedPercentageAllDays = sum(Visited_Matrix(:) ~= 0) / double(NumOfRowsToCheck*NumOfPlantsInRow)
*100 %percentage of visited plants
NumOfBugshaventFound =  sum(Bugs_Matrix(:) == 1) %Num Of Bugas that havent found yet

DetectionPercent (end+1) = numOfDetections/SumOfBugs * 100;
TimeDetectionPercent (end+1) = (totalTime*h - timeLeft)/60/60; %working time in Hours

plot(TimeDetectionPercent/2,DetectionPercent)
title('Detection Percent: Neighbor Strategy');
xlabel('Days');
ylabel('Percent detection of infected plants');
```

## A.7. Matlab Code- Random Strategy

```matlab
% initialize variables
LOR = 100; %Length of rows
WOR = 3.5; %Width of rows
days = 3; %number of days for the simulation
DistPlants = 2; %Distance Between Plants- meters
Dunam = 2 ;
Robot_Speed_KMH = 2.5; % Robot Speed - Kilometers per hour - average
Robot_Speed = (Robot_Speed_KMH*1000)/60/60; %Robot speed - Meters per Seconds
DiseaseProbability = 0.3; %chance of having a bug on the vine
DiseaseSeverity = 0.3; %defines how fast distributes to neighbor plants
NumOfPlantsInRow = int16(LOR/DistPlants); %calc number of plants in a single row
NumOfRows = int16(Dunam/(LOR*WOR)); %calc number of rows in the green house
NumOfRowsToCheck = int16(NumOfRows*2-2); %each row has two lines of plants- except of the first and the last.
oneVPTime = 4.34; %total seconds for one View Point
twoVPsTime = 16.14; %total seconds for two View Points
threeVPsTime = 23.74; %total seconds for three View Points
turningTime90 = 10; %total seconds for turning to the plants/back to road 90 degrees **Check the correct time**
turningTime180 = 23; %total seconds for turning 180 degrees  to the next
totalTime = 2*60*60; %total second robot works during the day
timeLeft = totalTime; %total seconds left for the rest of the day
timeBetweenPlants = DistPlants/Robot_Speed; %navigating time between two adjacent plants- seconds
timeBetweeenRows = 25; %navigating time between two adjacent rows- seconds
numOfDetections = 0; %count number of Correct Detections all days
numOfDetectionsToday = 0;%count number of Correct Detections this day
TotalBugsToday = 0;%sum number of bugs at the begining of a day
SumOfBugs = 0; %sum number of bugs
DetectionPercent = []; %List of detection percent - function of time
TimeDetectionPercent = []; %List of detection percent - function of time

Bugs_Matrix = zeros(NumOfPlantsInRow,NumOfRowsToCheck); %matrix for the bugs locations
Visited_Matrix = zeros(NumOfPlantsInRow,NumOfRowsToCheck);%presents which plants the robot visited
Visited_Today_Matrix = zeros(NumOfPlantsInRow,NumOfRowsToCheck);%presents which plants the robot visited Today

%adds Bugs to the matrix randomally
pos = 1; %index for the matrix cells
```

```matlab
for i = 1:(NumOfRowsToCheck/2+1)
    for j = 1:NumOfPlantsInRow
        if (i == (NumOfRowsToCheck/2 + 1))%last row
            if rand > (1-DiseaseProbability) %chances for having a bug on the vine
                Bugs_Matrix(pos - NumOfPlantsInRow) = 1;
            else
                Bugs_Matrix(pos - NumOfPlantsInRow) = 0;
            end
        elseif (i == 1) %first row
            if rand > (1-DiseaseProbability) %chances for having a bug on the vine
                Bugs_Matrix(pos) = 1;
            else
                Bugs_Matrix(pos) = 0;
            end
        else %all the rows except the first and the last
            if rand > (1-DiseaseProbability) %chances for having a bug on the vine
                Bugs_Matrix(pos) = 1;
                Bugs_Matrix(pos - NumOfPlantsInRow) = 1; %adjacent plant
            else
                Bugs_Matrix(pos) = 0;
            end
        end
        pos = pos +1;
    end
    pos = pos + NumOfPlantsInRow;
end
SumOfBugs = sum(Bugs_Matrix(:) == 1); %sum number of bugs

%strategy 3 - random sample - snake
n = randi(4);
VPtime = threeVPsTime;
pos = 1; %index for the matrix cells
forward = true; %indicates the direction the robot should go
for h = 1:days
    Visited_Today_Matrix = zeros(NumOfPlantsInRow,NumOfRowsToCheck);
    numOfDetectionsToday = 0;
    newBugsToday = 0;
    TotalBugsToday = sum(Bugs_Matrix(:) == 1);
    timeLeft = totalTime; %total seconds left for the rest of the day
    while((timeLeft-(timeBetweenPlants + VPtime + turningTime90)) > 0); % there is more working time
```

```matlab
if(forward == true)
  for i = 1:NumOfRowsToCheck;
    if(pos + n > (NumOfRowsToCheck*NumOfPlantsInRow))%out of bounds
      forward = false;
      break
    end
    for j = 1:n:NumOfPlantsInRow;
      n = randi(5);
      if((timeLeft-(timeBetweenPlants + VPtime)) > 0); %working time hasn't finished yet, and the robot can check another vine
        timeLeft = timeLeft - (timeBetweenPlants + VPtime);
        Visited_Matrix(pos) = Visited_Matrix(pos)+ 1;
        Visited_Today_Matrix(pos) = Visited_Today_Matrix(pos)+ 1;
        if (Bugs_Matrix(pos) == 1); %in case of detection
          Bugs_Matrix(pos) = 0;
          numOfDetections = numOfDetections + 1;
          numOfDetectionsToday = numOfDetectionsToday + 1;
        end
        if(j == NumOfPlantsInRow);%if it is the last vine in the row
          if(needToMoveBetweenRows == true)%
            timeLeft = timeLeft - timeBetweeenRows; %calculate time between rows
          else
            timeLeft = timeLeft - turningTime180; %calculate turning time at the same row
          end
        end
      else
        break
      end
      if(pos + n > (NumOfRowsToCheck*NumOfPlantsInRow))%out of bounds
        if(pos + 1 < (NumOfRowsToCheck*NumOfPlantsInRow))
          pos=pos+1;
        else
          pos=pos-1;
        end
        forward = false;
        break
      else
        pos = pos + n;
      end
    end
```

```matlab
        end
    else
        for i = NumOfRowsToCheck:-1:1;
            if(pos - n < 1) %out of bounds
                forward = true;
                break
            end
            for j = NumOfPlantsInRow:-n:1;
                n = randi(5);
                if((timeLeft-(timeBetweenPlants + VPtime)) > 0); %working time hasn't finished yet, and the robot can check another vine
                    timeLeft = timeLeft - (timeBetweenPlants + VPtime);
                    Visited_Matrix(pos) = Visited_Matrix(pos)+ 1;
                    Visited_Today_Matrix(pos) = Visited_Today_Matrix(pos)+ 1;
                    if (Bugs_Matrix(pos) == 1); %in case of detection
                        Bugs_Matrix(pos) = 0;
                        numOfDetections = numOfDetections + 1;
                        numOfDetectionsToday = numOfDetectionsToday + 1;
                    end
                    if(j == NumOfPlantsInRow);%if it is the last vine in the row
                        if(needToMoveBetweenRows == true)
                            timeLeft = timeLeft - timeBetweeenRows; %calculate time between rows
                        else
                            timeLeft = timeLeft - turningTime180; %calculate turning time at the same row
                        end
                    end
                else
                    break
                end
                if(pos - n < 1) %out of bounds
                    if(pos - 1 > 1)
                        pos=pos-1;
                    else
                        pos=pos+1;
                    end
                    forward = false;
                    break
                else
                    pos = pos - n;
                end
```

```matlab
            end
         end
      end
   end
   VisitedToday = sum(Visited_Today_Matrix(:) ~= 0) / double(NumOfRowsToCheck*NumOfPlantsInRow)*100;
   VisitedTodayToPrint = ['Visited Percentage of day number ',num2str(h), ' is: ' , num2str(VisitedToday)];
   disp (VisitedTodayToPrint)
   DetectionPercentageToday = numOfDetectionsToday/TotalBugsToday*100;
   DetectionPercentageToPrint = ['Detection Percentage of day number ',num2str(h), ' is: ' , num2str(DetectionPercentageToday)];
   disp(DetectionPercentageToPrint)
   pos_Bugs_Matrix = 1;
   Helper_Bugs_Matrix = Bugs_Matrix;
   rowIndex = 1 ; %index for which row
   left = true; % it has neighbor at the right side
   while (pos_Bugs_Matrix < NumOfPlantsInRow * NumOfRowsToCheck + 1)
      if (rowIndex == 1 ) % checking first row
         if(pos_Bugs_Matrix == 1 && Helper_Bugs_Matrix(pos_Bugs_Matrix) == 1) % first vine first row
            if rand < DiseaseSeverity %chance of spreading the disease
               Bugs_Matrix(pos_Bugs_Matrix + 1) = 1;
               newBugsToday = newBugsToday + 1;
            end
         elseif (pos_Bugs_Matrix == NumOfPlantsInRow && Helper_Bugs_Matrix(pos_Bugs_Matrix) == 1) %last vine first row
            if rand < DiseaseSeverity %chance of spreading the disease
               Bugs_Matrix(pos_Bugs_Matrix - 1) = 1;
               newBugsToday = newBugsToday + 1;
               rowIndex = 2;
            end
         else %first row
            if (Helper_Bugs_Matrix(pos_Bugs_Matrix) == 1)
               if rand < DiseaseSeverity %chance of spreading the disease
                  Bugs_Matrix(pos_Bugs_Matrix - 1) = 1;
                  Bugs_Matrix(pos_Bugs_Matrix + 1) = 1;
                  newBugsToday = newBugsToday + 2;
               end
            end
         end
      elseif (rowIndex == NumOfRowsToCheck ) % checking last row
```

```matlab
        if(pos_Bugs_Matrix == NumOfPlantsInRow * (NumOfRowsToCheck - 1 ) + 1 && Helper_Bugs_Matrix(pos_Bugs_Matrix) == 1) % first vine last row
            if rand < DiseaseSeverity && Bugs_Matrix(pos_Bugs_Matrix + 1) ~= 1 %chance of spreading the disease
                Bugs_Matrix(pos_Bugs_Matrix + 1) = 1;
                newBugsToday = newBugsToday + 1;
            end
        elseif (pos_Bugs_Matrix == NumOfPlantsInRow * NumOfRowsToCheck && Helper_Bugs_Matrix(pos_Bugs_Matrix) == 1) %last vine first row
            if rand < DiseaseSeverity && Bugs_Matrix(pos_Bugs_Matrix - 1) ~= 1 %chance of spreading the disease
                Bugs_Matrix(pos_Bugs_Matrix - 1) = 1;
                newBugsToday = newBugsToday + 1;
            end
        else %last row
            if (Helper_Bugs_Matrix(pos_Bugs_Matrix) == 1)
                if rand < DiseaseSeverity %chance of spreading the disease
                    if Bugs_Matrix(pos_Bugs_Matrix - 1) ~= 1
                        Bugs_Matrix(pos_Bugs_Matrix - 1) = 1;
                        newBugsToday = newBugsToday + 1;
                    end
                    if Bugs_Matrix(pos_Bugs_Matrix + 1) ~= 1
                        Bugs_Matrix(pos_Bugs_Matrix + 1) = 1;
                        newBugsToday = newBugsToday + 1;
                    end
                end
            end
        end
    else % rows with neighbor
        if(pos_Bugs_Matrix - (NumOfPlantsInRow * (rowIndex - 1)) == 1 && Helper_Bugs_Matrix(pos_Bugs_Matrix) == 1) % first vine in the row
            if rand < DiseaseSeverity %chance of spreading the disease
                if left == true
                    if Bugs_Matrix(pos_Bugs_Matrix + 1) ~= 1
                        Bugs_Matrix(pos_Bugs_Matrix + 1) = 1;  % neigbhor from the right
                        newBugsToday = newBugsToday + 1;
                    end
                    if Bugs_Matrix(pos_Bugs_Matrix + NumOfPlantsInRow) ~= 1
                        Bugs_Matrix(pos_Bugs_Matrix + NumOfPlantsInRow) = 1;
                        newBugsToday = newBugsToday + 1;
```

```matlab
                end
            else
                if Bugs_Matrix(pos_Bugs_Matrix + 1) ~= 1
                    Bugs_Matrix(pos_Bugs_Matrix + 1) = 1;  % neigbhor from the left
                    newBugsToday = newBugsToday + 1;
                end
                if Bugs_Matrix(pos_Bugs_Matrix - NumOfPlantsInRow) ~= 1
                    Bugs_Matrix(pos_Bugs_Matrix - NumOfPlantsInRow) = 1;
                    newBugsToday = newBugsToday + 1;
                end
            end
        end
    elseif(pos_Bugs_Matrix - (NumOfPlantsInRow * (rowIndex - 1)) == NumOfPlantsInRow) %last vine in the row
        if left == true
         left = false;
            if(Helper_Bugs_Matrix(pos_Bugs_Matrix) == 1 && rand < DiseaseSeverity)
                if Bugs_Matrix(pos_Bugs_Matrix - 1) ~= 1
                    Bugs_Matrix(pos_Bugs_Matrix - 1) = 1;  % neigbhor from the right
                    newBugsToday = newBugsToday + 1;
                end
                if Bugs_Matrix(pos_Bugs_Matrix + NumOfPlantsInRow) ~= 1
                    Bugs_Matrix(pos_Bugs_Matrix + NumOfPlantsInRow) = 1;
                    newBugsToday = newBugsToday + 1;
                end
            end
        else
         left = true;
         if(Helper_Bugs_Matrix(pos_Bugs_Matrix) == 1 && rand < DiseaseSeverity)
                if Bugs_Matrix(pos_Bugs_Matrix - 1) ~= 1
                    Bugs_Matrix(pos_Bugs_Matrix - 1) = 1;  % neigbhor from the left
                    newBugsToday = newBugsToday + 1;
                end
                if Bugs_Matrix(pos_Bugs_Matrix - NumOfPlantsInRow) ~= 1
                    Bugs_Matrix(pos_Bugs_Matrix - NumOfPlantsInRow) = 1;
                    newBugsToday = newBugsToday + 1;
                end
            end
        end
    else
```

```matlab
                if (rand < DiseaseSeverity && Helper_Bugs_Matrix(pos_Bugs_Matrix) == 1) %chance of spreading the disease
                    if left == true
                        if Bugs_Matrix(pos_Bugs_Matrix - 1) ~= 1
                            Bugs_Matrix(pos_Bugs_Matrix - 1) = 1;
                            newBugsToday = newBugsToday + 1;
                        end
                        if Bugs_Matrix(pos_Bugs_Matrix + 1) ~= 1
                            Bugs_Matrix(pos_Bugs_Matrix + 1) = 1;
                            newBugsToday = newBugsToday + 1;
                        end
                        if Bugs_Matrix(pos_Bugs_Matrix + NumOfPlantsInRow) ~= 1
                            Bugs_Matrix(pos_Bugs_Matrix + 1) = 1;
                            newBugsToday = newBugsToday + 1;
                        end
                    else
                        if Bugs_Matrix(pos_Bugs_Matrix - 1) ~= 1
                            Bugs_Matrix(pos_Bugs_Matrix - 1) = 1;
                            newBugsToday = newBugsToday + 1;
                        end
                        if Bugs_Matrix(pos_Bugs_Matrix + 1) ~= 1
                            Bugs_Matrix(pos_Bugs_Matrix + 1) = 1;
                        end
                        if Bugs_Matrix(pos_Bugs_Matrix - NumOfPlantsInRow) ~= 1
                            Bugs_Matrix(pos_Bugs_Matrix - NumOfPlantsInRow) = 1;
                        end
                    end
                end
            end
        end
        pos_Bugs_Matrix = pos_Bugs_Matrix +1;
    end
    if h~=days
        SumOfBugs = SumOfBugs + newBugsToday;
    end
end

%performance
clearSpots = sum(Bugs_Matrix(:) == 0);
PercentOfDetections = clearSpots/double(NumOfRowsToCheck*NumOfPlantsInRow)
```

VisitedPercentageAllDays = sum(Visited_Matrix(:) ~= 0) / double(NumOfRowsToCheck*NumOfPlantsInRow)

*100 %percentage of visited plants

NumOfBugshaventFound =  sum(Bugs_Matrix(:) == 1) %Num Of Bugas that havent found yet

# תקציר

חרקים מזיקים ומחלות בצמחים הם גורמים עיקריים להפחתה בייצור בתעשייה החקלאית, ועשויות להיות להם השפעות משמעותיות על המגזר החקלאי. איתור מוקדם ככל האפשר של מזיקים יכול לסייע בהגדלת תפוקת היבול ויעילות הייצור. מערכות ניטור רובוטיות רבות פותחו, אשר מאפשרות לאסוף נתונים ולספק הבנה טובה יותר של תהליכים סביבתיים. רובוט חקלאי יכול לאפשר זיהוי מדויק ומהיר של מזיקים, על ידי נסיעה בשדה באופן אוטונומי וניטור השטח בתוך השדה. עם זאת, במקרים רבים אי אפשר לדגום את כל הצמחים בשדה עקב מגבלות משאבים, כגון מגבלות כספיות, מגבלות של זמן או אנרגיה.

בתזה זו, פותחו ונותחו מספר אלגוריתמי דגימה כדי להתמודד עם האתגר של רובוט קרקע לניטור חקלאי שנועד לאתר חרקים בשדה שבו דגימה מלאה של כל הצמחים אינה ניתנת לביצוע עקב אילוצי משאבים.

שתי סיטואציות מרכזיות נחקרו באמצעות פיתוח מודלים של סימולציה אשר פותחו במיוחד במסגרת תזה זו: מצב שבו אין מידע א-פריורי על החרקים ומצב שבו אכן קיים מידע מוקדם על התפלגות החרקים בשטח.

במצב של חוסר במידע א-פריורי על השדה ו/או החרקים שבו, פותחו ונבדקו שבעה אלגוריתמים, שכל אחד מהם משתמש בגישה שונה לדגימת השדה ללא ידע מוקדם עליו. תוצאות המחקר הניבו תובנות חשובות. עבור שדות קטנים ורובוט עובד יחיד, דגימה יסודית היא חיונית ויש לצמצם את הצמחים אשר מדלגים עליהם עד לכדי המינימום האפשרי. שיעורי הזיהוי הטובים ביותר נצפו עם אלגוריתמים שמבקרים בכל צמח, ודילוג על צמחים הוביל לירידה בשיעורי ההצלחה. עבור שדות גדולים יותר, האלגוריתם אשר הציג את התוצאות הטובות ביותר הינו זה המיישם את אסטרטגיית השכן, המספק כיסוי יסודי מבלי לדלג על צמחים ומאפשר שליטה על חלק גדול יותר מהשדה בכל יום. לאלגוריתם זה יש את הפוטנציאל הגבוה ביותר להגיע ל-100% זיהוי, ללא קשר לגודל השדה.

במצב בו אכן קיים מידע מוקדם על מיקומי החרקים בשטח, אנו מציגים את הפיתוח והניתוח של אלגוריתמת דגימה דינמי, המנצל מידע בזמן אמת לתיעדוף דגימה בנקודות חשודות, איתור נקודות ריכוז מזיקים והתאמת תוכניות דגימה בהתאם. ביצועי האלגוריתם הושוו לשני אלגוריתמים קיימים באמצעות נתוני חרקים ממחקר קודם. ניתוחים גילו שהאלגוריתמת הדינמי עלה על האחרים בכל התרחישים שנבדקו, והגיע ל-100% זיהוי ב-3-5 ימים מוקדם יותר ביישום על שדות קטנים, וזיהה כ- 30%-50% יותר חרקים עבור שדות גדולים יותר. אחוזי הגילוי הגבוהים שלו בשדות קטנים – 100% עבור שדה קטן בגודל של 1 דונם - ירדו במידה מתונה עם הגדלת גודל השדה ל-80% עבור שדה של 10 דונם, ללא קשר לקצב התפשטות החרקים, שגם בקושי השפיע על זיהוי החרקים. הכפלת הזמן שהושקע בכל דגימת צמח שיפרה את התוצאות ב-30-50% בממוצע בשלבים המוקדמים של הסימולציה.

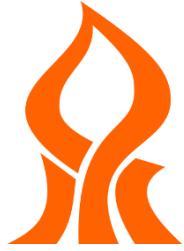

# אוניברסיטת בן-גוריון בנגב
## הפקולטה למדעי ההנדסה
המחלקה להנדסת תעשיה וניהול

# אלגוריתמי דגימה עבור רובוט נייד לניטור אוכלוסיות מזיקים בשדות חקלאיים

**חיבור זה מהווה חלק מהדרישות לקבלת תואר מגיסטר במדעי ההנדסה**

מאת: עדי יהושע
בהנחיית: פרופ' יעל אידן

| | |
|---|---|
| חתימת המחבר:............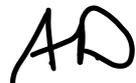........ | תאריך: 20.02.2023 |
| אישור המנחה:..................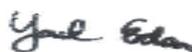............. ............ | תאריך: 20.02.2023 |
| אישור יו"ר ועדת תואר שני מחלקתי:................. | תאריך: |

שבט, תשפ"ג
מרץ, 2023

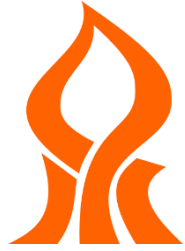

# אוניברסיטת בן-גוריון בנגב
## הפקולטה למדעי ההנדסה
המחלקה להנדסת תעשיה וניהול

# אלגוריתמי דגימה עבור רובוט נייד לניטור אוכלוסיות מזיקים בשדות חקלאיים

**חיבור זה מהווה חלק מהדרישות לקבלת תואר מגיסטר במדעי ההנדסה**

מאת: עדי יהושע
בהנחיית: פרופ' יעל אידן

שבט, תשפ"ג
מרץ, 2023